\documentclass{article}

\usepackage{arxiv}

\usepackage[utf8]{inputenc} 
\usepackage[T1]{fontenc}    
\usepackage{hyperref}       
\usepackage{url}            
\usepackage{booktabs}       
\usepackage{amsfonts}       
\usepackage{nicefrac}       
\usepackage{microtype}      
\usepackage{lipsum}		
\usepackage{graphicx}
\usepackage{natbib}
\usepackage{doi}
\usepackage[utf8]{inputenc}
\usepackage{subfig}
\usepackage{multirow}
\usepackage{amsmath}
\usepackage{changepage}
\usepackage{algorithm,algorithmic}
\usepackage[T1]{fontenc}
\usepackage{soul}
\newcommand{\orcid}[1]{\href{https://orcid.org/#1}{\includegraphics[scale=0.7]{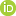}}}

\title{Ensemble of CNN classifiers using Sugeno Fuzzy Integral Technique for Cervical Cytology Image Classification}

\author{\href{https://orcid.org/0000-0000-0000-0000}{\includegraphics[scale=0.06]{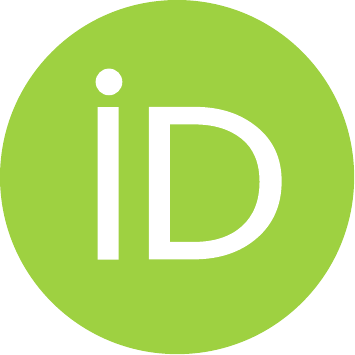} Rohit Kundu} \\
	Department of Electrical Engineering, \\Jadavpur University\\ Kolkata-700032, INDIA \\
	\And
	\href{https://orcid.org/0000-0000-0000-0000}{\includegraphics[scale=0.06]{orcid.pdf} Hritam Basak} \\
	Department of Electrical Engineering, \\Jadavpur University\\ Kolkata-700032, INDIA \\
	\And
	\href{https://orcid.org/0000-0000-0000-0000}{\includegraphics[scale=0.06]{orcid.pdf} Akhil Koilada} \\
	Department of Electrical Engineering, \\Jadavpur University\\ Kolkata-700032, INDIA \\
	\And
	\href{https://orcid.org/0000-0000-0000-0000}{\includegraphics[scale=0.06]{orcid.pdf} Soham Chattopadhyay} \\
	Department of Electrical Engineering, \\Jadavpur University\\ Kolkata-700032, INDIA \\	\And
	\href{https://orcid.org/0000-0000-0000-0000}{\includegraphics[scale=0.06]{orcid.pdf}Sukanta Chakraborty} \\
   Theism Medical Diagnostics Centre,\\ Kolkata-700030, INDIA \\
		\And
	\href{https://orcid.org/0000-0002-2426-9915}{\includegraphics[scale=0.06]{orcid.pdf}\hspace{1mm}Nibaran Das} \\
 Department of Computer Science and Engineering, \\Jadavpur University\\
  Kolkata-700032, INDIA\\
	\texttt{nibaran@ieee.org} \\ \thanks{
This work is partially supported by SERB GOI,( Ref. no. EEQ/2018/000963).}
}

\renewcommand{\shorttitle}{Sugeno Fuzzy Ensemble}

\hypersetup{
pdftitle={A template for the arxiv style},
pdfsubject={q-bio.NC, q-bio.QM},
pdfauthor={David S.~Hippocampus, Elias D.~Striatum},
pdfkeywords={First keyword, Second keyword, More},
}

\begin{document}
\maketitle

\begin{abstract}
	Cervical cancer is the fourth most common category of cancer, affecting more than 500,000 women annually, owing to the slow detection procedure. Early diagnosis can help in treating and even curing cancer, but the tedious, time-consuming testing process makes it impossible to conduct population-wise screening. To aid the pathologists in efficient and reliable detection, in this paper, we propose a fully automated computer-aided diagnosis tool for classifying single-cell and slide images of cervical cancer. The main concern in developing an automatic detection tool for biomedical image classification is the low availability of publicly accessible data. Ensemble Learning is a popular approach for image classification, but simplistic approaches that leverage pre-determined weights to classifiers fail to perform satisfactorily. In this research, we use the Sugeno Fuzzy Integral to ensemble the decision scores from three popular pretrained deep learning models, namely, Inception v3, DenseNet-161 and ResNet-34. The proposed Fuzzy fusion is capable of taking into consideration the confidence scores of the classifiers for each sample, and thus adaptively changing the importance given to each classifier, capturing the complementary information supplied by each, thus leading to superior classification performance. We evaluated the proposed method on three publicly available datasets, the Mendeley Liquid Based Cytology (LBC) dataset, the SIPaKMeD Whole Slide Image (WSI) dataset, and the SIPaKMeD Single Cell Image (SCI) dataset, and the results thus yielded are promising. Analysis of the approach using GradCAM-based visual representations and statistical tests, and comparison of the method with existing and baseline models in literature justify the efficacy of the approach.
\end{abstract}

\keywords{First keyword \and Second keyword \and More}

\section{Introduction}\label{intro}
One of the most prevalent and deadliest diseases of the 21st century is cancer which is caused by the uncontrolled growth of human body cells. Statistically, this is the second leading cause of death worldwide, causing around 9.6 million deaths every year and about 1/6th of the total deaths of the population throughout the globe. Late diagnosis and prognosis of cervical cancer often lead to deaths without receiving adequate treatment, mostly in poor and middle-income countries where the living index is low and healthcare infrastructure is insufficient.

Papanicolaou smear or the Pap smear test is the most common and widely used method in cervical cytology for the screening of abnormal lesions and cervical cancer. However, the assessment of cervical cytology requires expert physicians, which is expensive and time-consuming. Also, inter and intra-human variability of assessment or incorrect prognosis due to human errors may worsen the patient condition and can even be fatal in some cases. To avoid ambiguity in diagnosis, people tend to take the opinions of multiple experts. In the present work, we have utilized that kind of strategy for the automated detection of cervical cancer, where multiple CNN classifiers have been used to generate predictions, and the decision scores have been ensembled to conclude the final predictions.

\begin{figure*}
    \centering
    \includegraphics[scale=0.5]{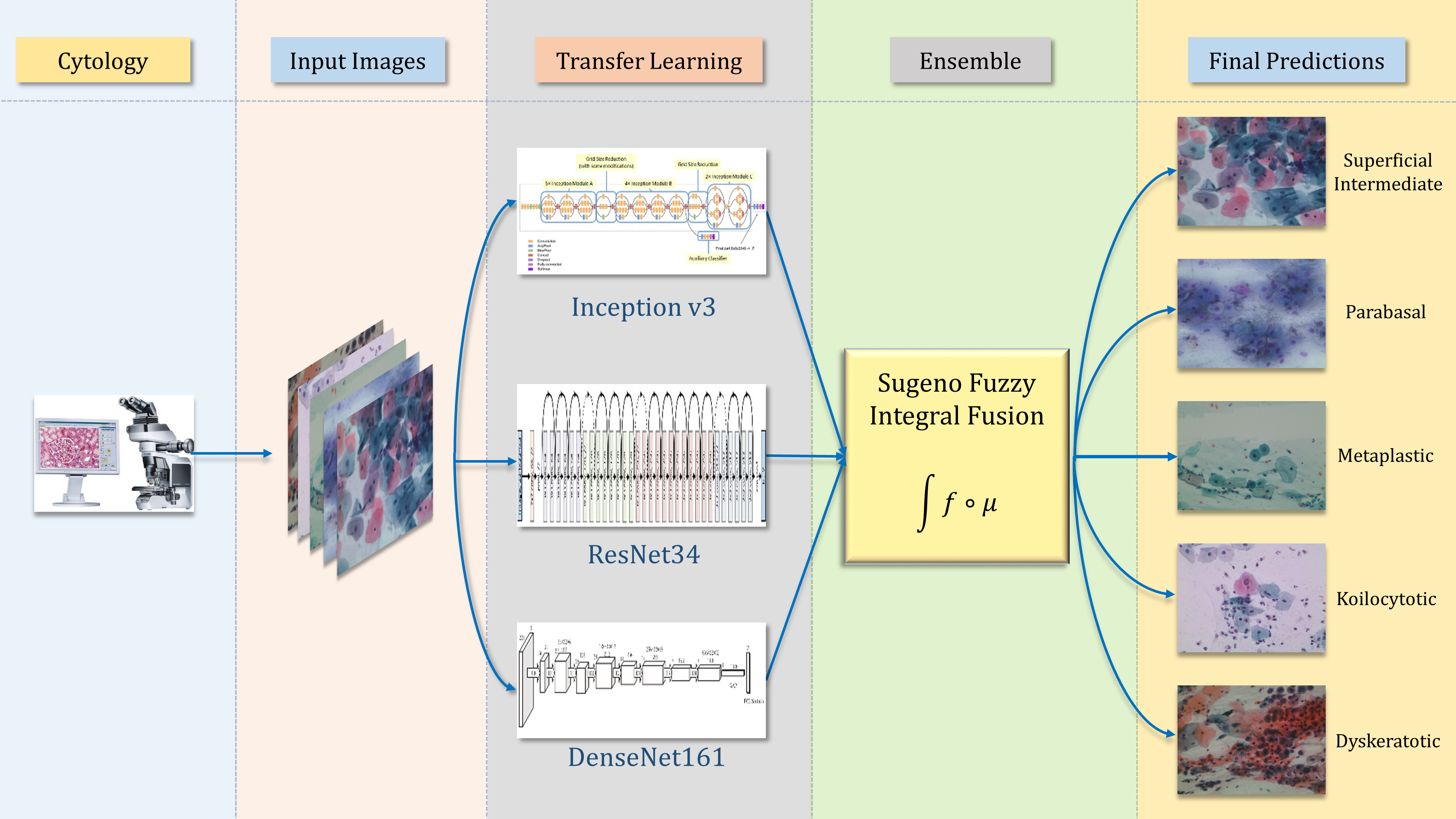}
    \caption{Overall workflow of the proposed framework}
    \label{overall}
\end{figure*}

In literature, there are some simple fusion schemes \cite{nannia2020ensemble}, that combines CNN features mostly by late fusion of several feature sets, specifically, by using majority voting, weighted probability ensemble, etc. These schemes utilize different CNN features by simple addition, multiplication or averaging them. Besides, experiments with different fusion method reveal some of the optimum weights that have been used for a weighted mean of the decision scores obtained. Therefore, there remains an opportunity to optimize the fusion schemes of different CNN or machine learning-based classifiers by adaptively promoting the importance of every classifier for every single image. This can be done by conditioning the weightage of one classifier upon others before it, which is done in a fuzzy fusion method, which remains largely unexplored in the particular domain of cervical cytology. The proposed method performed superior to other popularly used fusion methods as described in Section \ref{results}.

The rest of the paper is organised as follows: Section \ref{survey} provides a brief literature survey of the existing classification approaches and fusion methods in the domain of cervical cytology; Section \ref{proposedmethod} is the detailed description of the proposed method, where we have implemented transfer learning-based decision score generation followed by fuzzy fusion; Section \ref{results} contains information about the datasets used and the results we have obtained along with the comparison of different existing methods; Section \ref{conclusions} is the brief description of the outcome of our experiments and the future improvements that can be made to enhance the classification performance further.
\subsection{Overview and Contributions}
The high rate of cervical cancer cases in the world, especially in developing and underdeveloped countries is mainly due to inadequate screening. However, detection of cervical cancer is no easy feat, taking long hours to detect a single case, thus making regular population-wide screening an impossible task. This calls for the need for automation in the detection procedure, and thus in this paper, we propose a framework for reliable automated detection of cervical cancer employing deep neural networks and Ensemble Learning using Fuzzy Fusion. The overview of the proposed framework is shown in \autoref{overall}.

The contributions of this paper are as follows:
\begin{enumerate}
    \item The Sugeno Fuzzy Integral is introduced for the first time for cervical cell classification to fuse the decision scores of multiple CNN classifiers. Using it the performance of three popular individual CNN classifiers such as Inception v3 \cite{szegedy2016rethinking}, ResNet-34 \cite{he2016deep} and DenseNet-161 \cite{huang2017densely} are improved using the ensemble technique and thus has been used in the present research to make accurate predictions on the small-sized available datasets.  We have performed experiments with several ensemble approaches, but the Sugeno Fuzzy Integral-based ensemble outperformed the traditional methods \cite{nannia2020ensemble} since it is capable of using adaptive weights based on the confidence of predictions by the individual classifiers for each test sample, leading to superior predictions.
    \item The proposed method has been tested on two publicly available datasets: the SIPaKMeD Pap Smear image dataset \cite{plissiti2018sipakmed} and the Mendeley Liquid Based Cytology dataset \cite{hussain2020liquid}. The SIPaKMeD dataset has both whole slide images (WSI) and single-cell images (SCI), and hence both types have been used separately for evaluation. Promising results have been obtained by the framework, which is reliable for practical use.
\end{enumerate}

Thus we have developed an automated framework for the classification of Pap stained and Liquid-Based Cytology images using Deep Learning and a novel ensemble approach for the classification of cervical cytology images, which is otherwise a laborious task for cyto-technicians.
\begin{figure*}
    \centering
    \subfloat[Mendeley]{\includegraphics[scale=0.3]{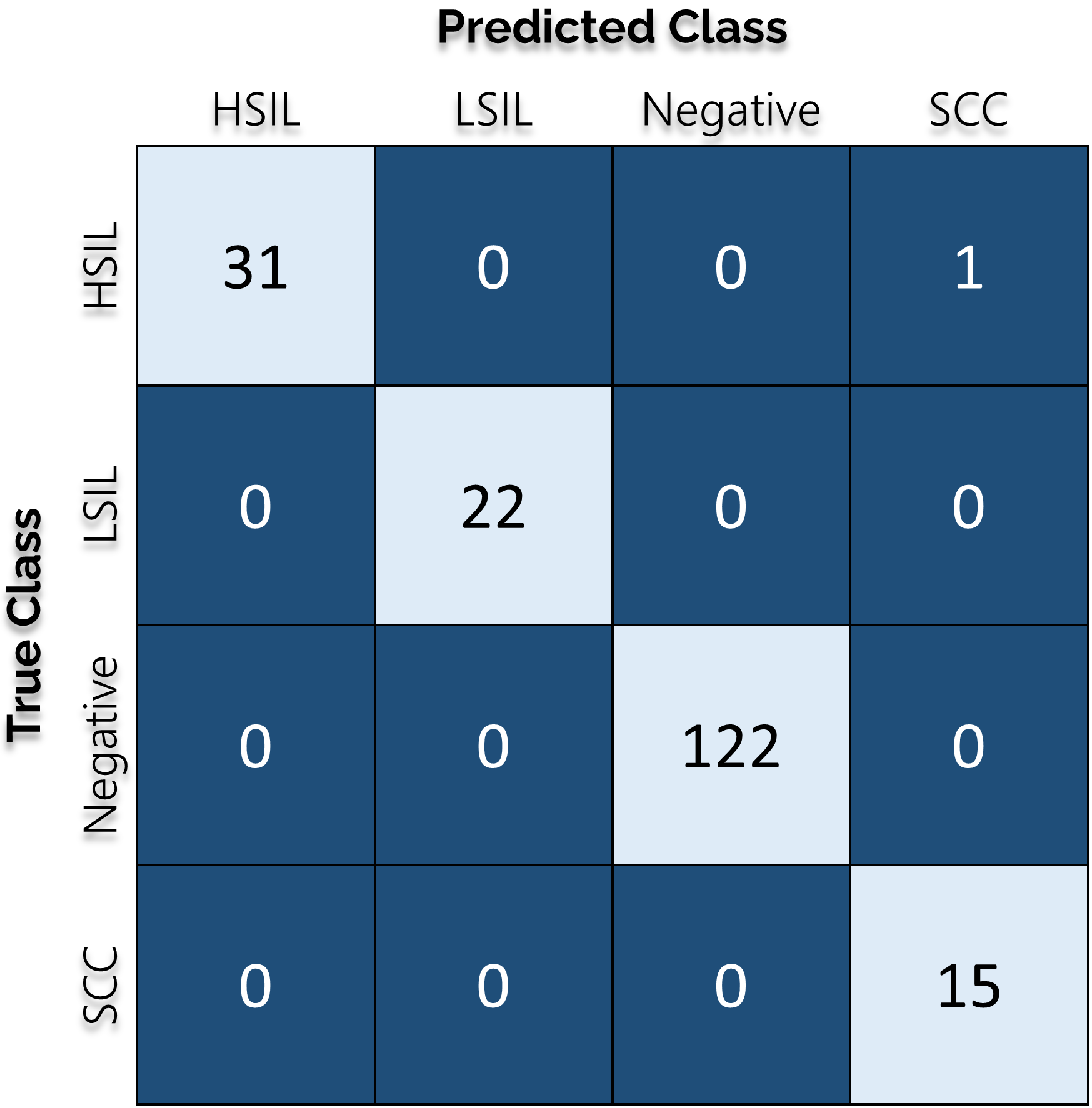}}\;\;
    \subfloat[SIPaKMeD WSI]{\includegraphics[scale=0.3]{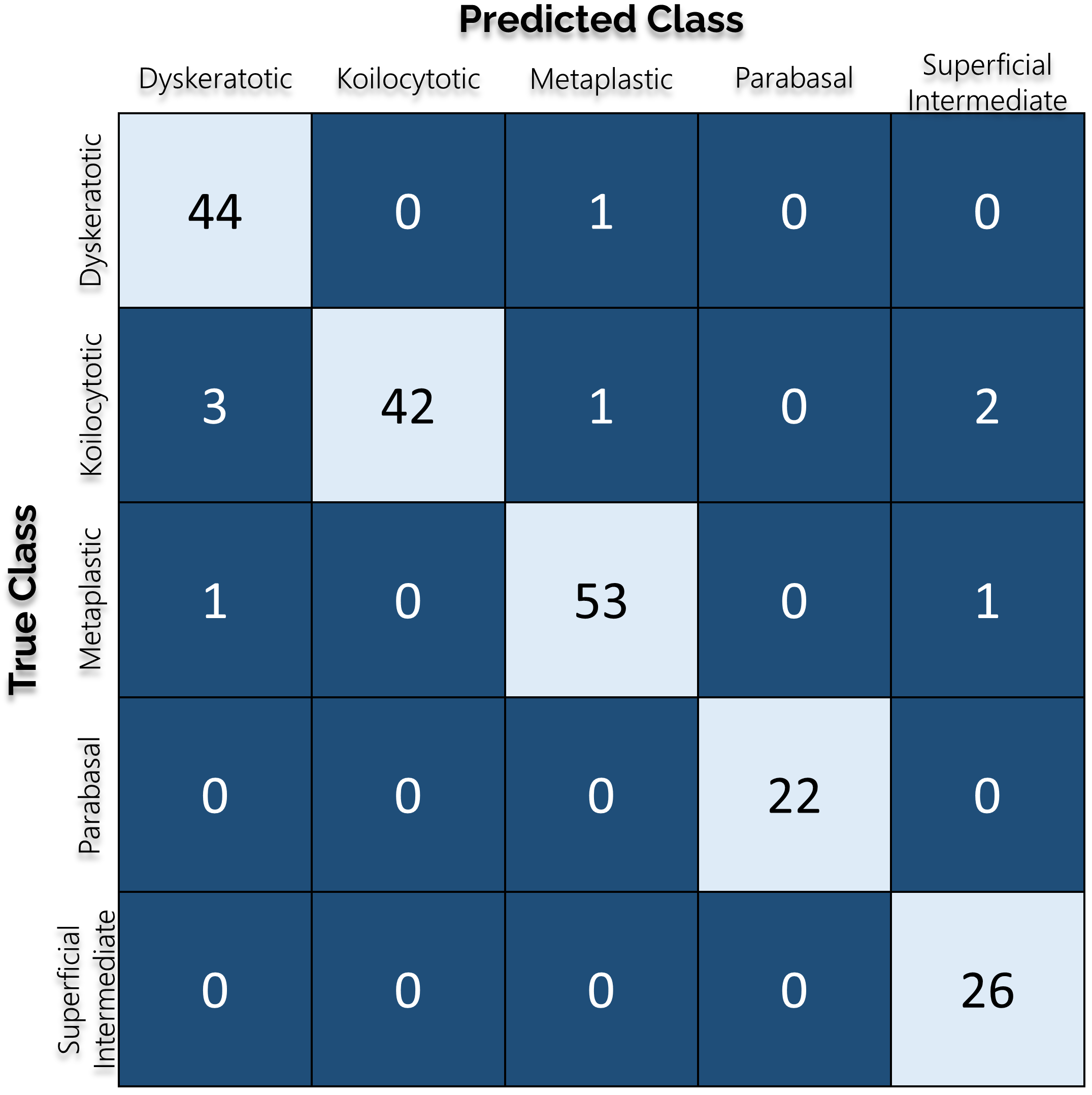}}\;\;
    \subfloat[SIPaKMeD SCI]{\includegraphics[scale=0.3]{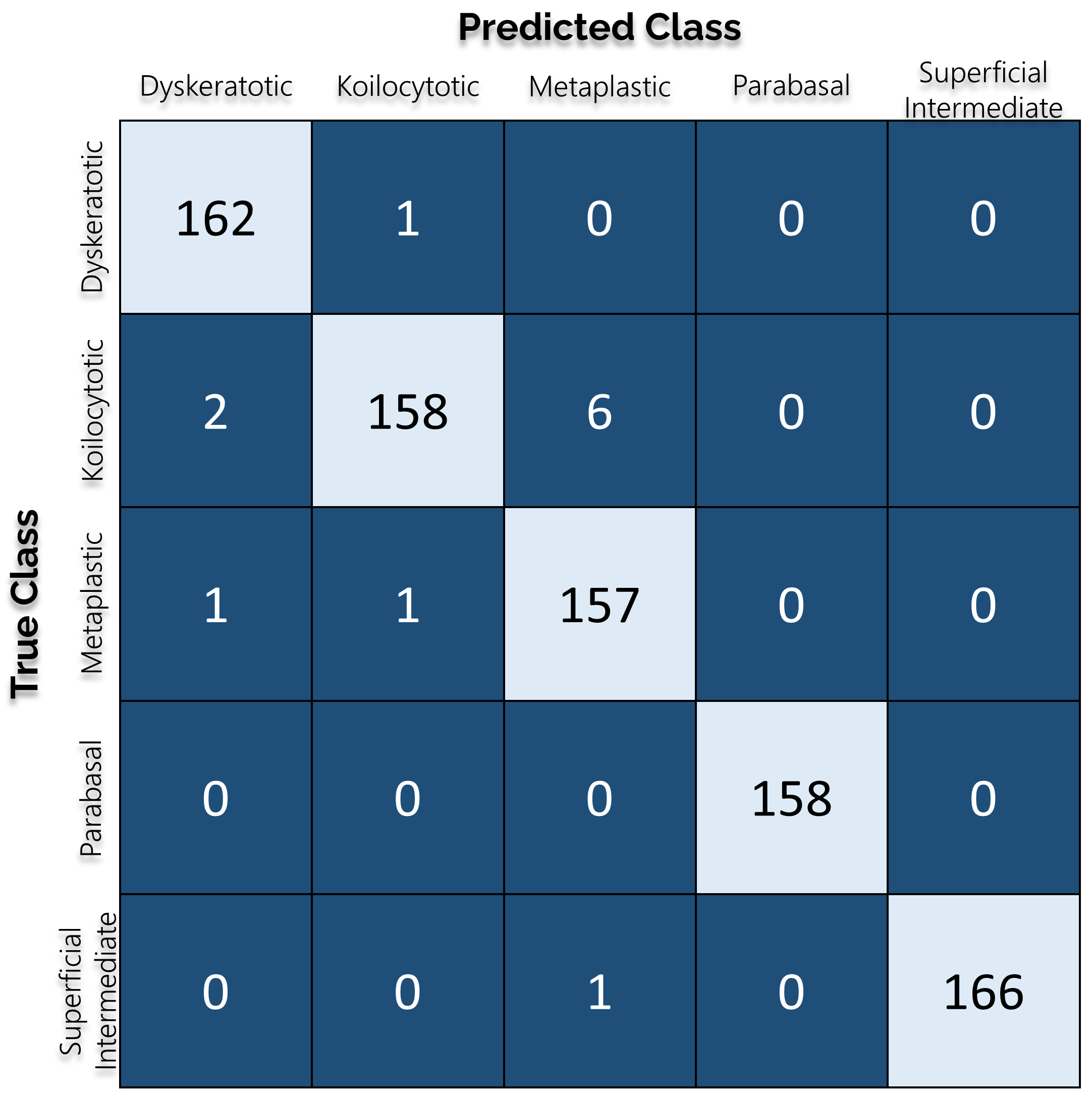}}
    \caption{Confusion Matrices of the respective test sets of (a) Mendeley LBC Dataset (b) SIPaKMeD Whole Slide Images Dataset (c) SIPaKMeD Single Cell Images Dataset}
    \label{cm}
\end{figure*}
\section{Literature Survey}\label{survey}
For many decades extensive researches have been done to develop improved algorithms and methods for computer-aided diagnosis of cervical cancer \cite{mitra2021cytology}. In the past few years, various machine learning algorithms have been proposed for the detection and classification of cancerous cell images such as Support Vector Machine used by Ashok et al. \cite{ashok2016comparison} and K-Nearest Neighbour classifier used by \cite{sharma2016classification}, etc.


Win et al. \cite{win2020computer} proposed a method in which nuclei were detected using a shape-based iterative method, and the overlapping cytoplasm was separated by a marker-control watershed approach. Features were extracted from regions of segmented nuclei and a Random Forest classifier was used for feature selection. For classification, bagging ensemble classifier, which combined the results of LDA, SVM, KNN, boosted trees and bagged trees. They achieved 98.27\% accuracy in two-class and 94.09\% accuracy in five-class classification on the SIPaKMeD dataset. Jia et al. \cite{jia2020detection} proposed a new framework based on a strong feature Convolutional Neural Networks-Support Vector Machine (CNN-SVM) model to classify cervical cells.  A method fusing the strong features extracted by Gray-Level Co-occurrence Matrix and Gabor filters with abstract features from the hidden layers of CNN was conducted, meanwhile, the fused ones were input into the SVM for classification. Basak et al. \cite{basak2021cervical} used a deep learning-based method where in they extracted deep features from multiple CNN models and applied a two-step feature enhancement procedure using Principle Component Analysis (PCA) and Grey Wolf Optimizer (GWO) to reduce the dimensionality of the feature set for efficient classification.

Ensemble Learning is a popular technique to incorporate the salient features of multiple CNN models like Kuko et al. \cite{kuko2019ensemble} proposed a method of applying Random Forest classifiers on another layer of ensemble learning based on the rotation of the image. Each image is rotated 8 times by 45 degrees and 8 Random Forest models were trained. After the classifications, an ensemble voting technique is used to tally all votes amongst the 8 models and the most-voted class is selected as the final classification. This method achieved an accuracy of 90.37\% on binary classification. Xue et al. \cite{xue2020application} used an ensemble learning method the weighted voting based method. They have developed Inception-V3, Xception, VGG-16, and ResNet-50 based TL structures. Then, to enhance the classification performance, a weighted voting based EL strategy was introduced. An experiment for classifying the benign cells from the malignant ones is carried out on the Herlev dataset and obtains an overall accuracy of 98.37\%. Sarwar et al. \cite{sarwar2015hybrid} used an ensemble system developed using Random subset space, Radial basis function network, Multiclass classifier, Random forest, Bagging, Rotation Forest, J48 graft, Ensemble of Nested dichotomies (END). Decorate, PART, Random Committee, Filtered Classifier, Decision Table, Multiple back propagation artificial neural network, and Naïve Bayes. The final classification decision is obtained by aggregating the output of all possible candidate trees for the multiclass problem. The overall accuracy of the system for the two-class problem was 98.57\% on the HErlev dataset.
\section{Proposed Method}\label{proposedmethod}
In the present study, we have used three different CNN architectures (Transfer Learning) for generating the confidence scores on the datasets: Inception v3, DenseNet-161 and ResNet-34. The decision scores from these classifiers have been fused using the Sugeno Fuzzy Integral to generate the final predictions. These steps are explained in detail in this section.

\subsection{Inception v3}
One of the most popularly used deep learning network for transfer learning technique is Inception v3 \cite{szegedy2016rethinking}, which is consisted of several inception blocks. It takes an input image of size $299\times299\times3$ and produces feature maps of different dimensions in different layers. The inception block of Inception v3 allows us to utilize the facilities of using different filters of feature extraction from a single feature map. These features with different filters are concatenated and passed on to the next layer for deeper feature extraction. In this study, we have evaluated Inception v3 with the ReLU activation function. In each case, the model is trained for 100 epochs with cross-entropy loss which is optimized by an SGD optimizer with a learning rate of 0.001. 
\subsection{DenseNet-161}
DenseNet has been proposed by \cite{huang2017densely} to address the problem of gradient vanishing for the case of deep neural networks. The building blocks of DenseNets are connected densely to each other. In this way, only fewer parameters are needed to be learnt by the network. These kinds of networks have very narrow architecture and add small sets of feature maps. This network also takes input image of size $224\times224\times3$ and similar to Inception v3, in our study, we have trained this model for 200 epochs and SGD optimizer with a 0.001 learning rate. For DenseNet-161 also we have used the ReLU activation function. 
\subsection{ResNet-34}
ResNet \cite{he2016deep} is also an advanced convolutional neural net with residual skip connection embedded in it. There are certain versions of ResNets, which are ResNet-18, ResNet-34, ResNet-50 and ResNet-152. Due to the embedding of the skip-connections, despite having such deep architecture, the gradient vanishing problem is already being taken care of. Similar to DenseNet, the standard image dimension which should be given as an input to any version of ResNet is $224\times224\times3$. We have evaluated ResNet-34 in this study. To maintain consistency, the number of epochs in training, the optimizer, learning rate etc. have been fixed to the values mentioned in the above CNNs.  
\subsection{Ensemble: Sugeno Fuzzy Integral}
To leverage the ascendency of individual CNN classifiers instead of a single one, we propose an integration of multiple classifiers utilizing fuzzy fusion in this paper.  As shown in \autoref{overall}, the confidence scores from multiple classifiers are treated as the input of the fuzzy fusion directly. It has been used previously in a pattern recognition task, specifically in classifier fusion \cite{wu2016fuzzy, liu2009machinery}, and has shown promising results. However, no such applications in the domain of cervical cytology have been found so far. The fusion scheme harnesses additional information of a classifier, which is the uncertainty of the decision scores. The generalization of aggregation operators for a set of confidence values is known as fuzzy measures, that uses some weights before each source. 

If $S = {S_k}_{k=1:N}$ be the set of $N$ classifiers, the fuzzy measure is the worth value of the set $S_k$ and, as introduced in \cite{sugeno1993fuzzy}, can have values in the range of [0, 1] and can be represented by the function $g(S_k)$. $g(S)=1$ represents that the classifier can be considered as consistent whereas $g(phi)=0$ represents that the classifier cannot be trusted and considered as the results are inconsistent. For all $c\subset S$, the fuzzy measure can be characterized by the monotonic property as in \autoref{mon_prop}.

\begin{equation}\label{mon_prop}
    c_a\subset c_b \:\:\Longrightarrow \:\: g(c_a)\subset g(c_b)
\end{equation}

Fuzzy density is defined as the fuzzy measure of set S when S contains a single element and is the measure of the worth value of individual classifiers. Some studies have used fuzzy density values predefined based on the experience of the researcher; however, that does not ensure superior integration of the classifiers. Instead, following the original work of \cite{tong2016speech}, we have set the fuzzy density values the same as the classifier accuracy measure on the test set, to give weightage to the optimal classifiers and punish the inferior ones.  Following the work of \cite{tahani1990information}, Sugeno fuzzy measure $\lambda$ can be conceptualized with an additional characteristic that if $c_a\cap c_b = \phi$ then it can be considered that there is always a $\lambda > -1$ such that:
\begin{equation}
g(c_a\cup c_b)= g(c_a)+g(c_b)+ g(c_a)\cdot g(c_b)\cdot \lambda,\:\:\ c_a,c_b\subset S
\end{equation}
The $\lambda$ value can be obtained by solving the following equation, where $\lambda$ is greater than -1.
\begin{equation}
\lambda+1=\prod_{i=1}^N \left(g(c_i)\cdot\lambda+1\right)
\end{equation} 
where $N$ is the number of CNN classifier, which is 3 in our case. $\lambda\in(-1,\infty)$ has the following characteristics:
\begin{enumerate}
\item $\lambda=0$ when $\sum_{i=1}^N g(c_i)=1$
\item $\lambda>0$ when $\sum_{i=1}^N g(c_i)<1$ 
\item $-1\leq\lambda<0$ when $\sum_{i=1}^N g(c_i)>1$
\end{enumerate}
Among all the existing methods of fuzzy integrals like Fuzzy min-max \cite{mesiar2008fuzzy}, ordered weighted averaging operators like ordered weighted averaging OR (OWA-OR) \cite{cheng2012combining} and ordered weighted averaging and (OWA-AND) \cite{cho1995fuzzy}, Sugeno integral \cite{sugeno1993fuzzy}, Choquet integral \cite{murofushi1989interpretation}, we have implemented Sugeno and Choquet integrals in this work, and have selected the best result from these two. The steps for calculating the fuzzy integrals are described below.

First, the N classifiers are sorted according to their output scores:
\begin{equation}
S_{\phi}^1\geq S_{\phi}^2\geq S_{\phi}^3\geq ....S_{\phi}^j\geq ....S_{\phi}^N
\end{equation}

where $S_{\phi}^j$ represents the $j^{th}$ largest output value of the classifiers where $j\in [1, N]$, where N is the number of classifiers. 
Next, we calculate the Choquet and Sugeno fuzzy integrals by means of:
\begin{equation}
Y_{\text{Choquet}}=\sum_{k=1}^N S_{\phi}^k \left( g(X_k)-g(X_{k-1}) \right)
\end{equation} 
 \begin{equation}
Y_{\text{Sugeno}}=\bigvee _{k=1}^N \left( S_{\phi}^k \wedge g(X_k) \right)
\end{equation}

where $X_k$ is defined from the definition of Sugeno fuzzy measures as follows:
\begin{equation}
X_k=X_{k-1}+\{S_{\phi}^k\}; \:\: X_0=\text{\{null\}}
\end{equation}
Thus through fuzzy integrals, the robustness is experimentally found to be higher as compared to the previously obtained normalized softmax probabilities and the time complexity of the algorithm is found to be $O(N\:log(N)\times P)$, where $N$ is the number of classifiers and $P$ are the number of classes. The pseudo-code for computing the Sugeno Fuzzy Integral for the ensemble of CNN classifiers decisions is shown in Algorithm \ref{sugalgo}.

\begin{algorithm}[tbp]
{\small
{\em Input:} \\
Set of Decision Scores (from different base learners): $D$\\
Set of Fuzzy Measures: $\mu$\\
{\em Output:}\\
Final Predictions on test samples: $\hat{y}$\newline
\begin{algorithmic}[] 
    \STATE predictions $\longleftarrow$ Initialize empty list of final predictions
    \STATE \FOR{class index  $c \in$ num\_classes}
        \STATE $D_{\pi}\leftarrow$ Permutation of $D$ in descending order
        \STATE $\mu_{\pi} \leftarrow$ Permutation of $\mu$ corresponding to $D_{\pi}$
        \STATE $f\textsubscript{prev} \leftarrow \mu_{\pi}[0]$
        \STATE pred $\leftarrow D_{\pi}[0] \times \mu_{\pi}[0]$
        \STATE \FOR{$n \in \{1,2,3,\hdots,N\}$}
            \STATE $f\textsubscript{current} \leftarrow f\textsubscript{prev}+\mu_{\pi}[N]+ \lambda.\mu_{\pi}[N]\times f\textsubscript{prev}$
            \STATE pred $\leftarrow$ pred$+D_{\pi}\times(f_{current} - f_{prev})$
            \STATE $f_{prev} \leftarrow f_{current}$
        \ENDFOR
        \STATE predictions[c] $\leftarrow$ pred
    \ENDFOR
    \STATE $\hat{y} \leftarrow argmax(predictions[c])$
\end{algorithmic}
\caption{Ensemble of Classifiers Decisions using Sugeno Fuzzy Integral }
\label{sugalgo}
}
\end{algorithm}

\section{Results and Discussion}\label{results}
In this section, we first briefly describe the two publicly available datasets used. Then we evaluate the performance of the proposed framework on these datasets and compare the results to other popular approaches used in literature to justify the viability of the method used.
\subsection{Description of Datasets}
In the present study, we have used two publicly accessible datasets: the SIPaKMeD Pap Smear dataset by Plissiti et al \cite{plissiti2018sipakmed} and the Mendeley Liquid Based Cytology (LBC) dataset by Hussain et al. \cite{hussain2020liquid} which are briefly described below.
\subsubsection{SIPaKMeD Pap Smear Dataset}
\begin{table}[]
\centering
\caption{SIPaKMeD Pap Smear Dataset Distribution. WSI: Whole Slide Images, SCI: Single Cell Images}
\label{sipak}
\begin{tabular}{|c|c|c|c|}
\hline
\textbf{Class} & \textbf{Category}        & \textbf{WSI} & \textbf{SCI}  \\ \hline
1              & Superficial-Intermediate & 126          & 831           \\ \hline
2              & Parabasal                & 108          & 787           \\ \hline
3              & Koilocytotic             & 238          & 825           \\ \hline
4              & Metaplastic              & 271          & 793           \\ \hline
5              & Dyskeratotic             & 223          & 813           \\ \hline
-              & \textbf{Total}           & \textbf{966} & \textbf{4049} \\ \hline
\end{tabular}
\end{table}
The SIPaKMeD dataset consists of 4049 images of isolated cells that have been manually cropped from 966 cluster cell images of Pap smear slides, which are also included. The cells are classified into five different classes by expert cytopathologists. Normal cells are divided into two categories (superficial-intermediate, parabasal), abnormal but not malignant cells are divided into two categories (koilocytes and dyskeratotic) and the final category is benign (metaplastic) cells. Both the Whole Slide Images (WSI) and the Single Cell Images (SCI) have been used separately for the present study. The distribution of images in the dataset is shown in \autoref{sipak}.
\subsubsection{Mendeley Liquid Based Cytology Dataset}
\begin{table}[]
\centering
\caption{Mendeley LBC Dataset Distribution. NILM: Negative for Intra-epithelial Malignancy, HSIL: High Squamous Intra-epithelial Lesion, LSIL:  Low Squamous Intra-epithelial Lesion, SCC: Squamous Cell Carcinoma}
\label{mendeley}
\begin{tabular}{|c|c|c|}
\hline
\textbf{Class} & \textbf{Category} & \textbf{Number of Images}   \\ \hline
1              & NILM              & 613                         \\ \hline
2              & HSIL              & 113                         \\ \hline
3              & LSIL              & 163                         \\ \hline
4              & SCC               & 74                          \\ \hline
-              & \textbf{Total}    & \textbf{963}                \\ \hline
\end{tabular}
\end{table}
The Liquid Based Cytology (LBC) Dataset by Hussain et al. \cite{hussain2020liquid} contains 963 images classified into four different classes. The Pap smear images were captured in 40x magnification which is collected and prepared using the liquid-based cytology technique from 460 patients. In this dataset 613 images belong to the normal cell category and 350 images belong to the abnormal cell category. The distribution of these images is given in \autoref{mendeley}.
\subsection{Metrics for Performance Evaluation}
For our work, we have used Accuracy, Precision, Recall and F1-score for evaluating the performance of the proposed framework. True Positive (TP), True Negative (TN), False Positive (FP) and False Negative (FN) are the basic elements that help determine the values of these metrics, and they can be defined as follows:
\begin{itemize}
\item True Positive: The predicted result is positive, while it is labelled as positive.
\item False Positive: The predicted result is positive, while it is labelled as negative. It calls Type I Error as well.
\item True Negative: The predicted result is negative, while it is labelled as negative.
\item False Negative: The predicted result is negative, while it is labelled as positive. It calls Type II Error as well.
\end{itemize}
Based on these 4 elements, we can calculate the metrics: accuracy, precision, recall, F1 score. For a multi-class system ($N$ class), if we have a confusion matrix $M$, with the rows depicting the predicted class, and the columns depicting the true class, these evaluations metrics can be formulated as in Equations \ref{acc}, \ref{pre}, \ref{rec} and \ref{f1}.
\begin{equation}
Accuracy = \frac{\sum_{i=1}^{N}M\textsubscript{ii}}{\sum_{i=1}^{N}\sum_{j=1}^{N}M\textsubscript{ij}}
\label{acc}
\end{equation}
\begin{equation}
Precision\textsubscript{i} = \frac{M\textsubscript{ii}}{\sum_{j=1}^{N}M\textsubscript{ji}}
\label{pre}
\end{equation}
\begin{equation}
Recall\textsubscript{i} = \frac{M\textsubscript{ii}}{\sum_{j=1}^{N}M\textsubscript{ij}}
\label{rec}
\end{equation}
\begin{equation}
F1-Score\textsubscript{i} = \frac{2}{\frac{1}{Precision\textsubscript{i}}+\frac{1}{Recall\textsubscript{i}}}
\label{f1}
\end{equation}
\subsection{Implementation}
\begin{table*}[]
\centering
\caption{Class-wise and Net Results obtained on the three datasets}
\label{res_data}
\begin{tabular}{|c|c|c|c|c|c|}
\hline
\textbf{Dataset} & \textbf{Class} & \textbf{Precision(\%)} & \textbf{Recall(\%)} & \textbf{F1 Score(\%)} & \textbf{Accuracy(\%)} \\ \hline
\multirow{5}{*}{\textbf{Mendeley LBC}} & High Squamous Intra-epithelial Lesion             & 100            & 93.75          & 96.77          & 93.75          \\ \cline{2-6} 
                                       & Low Squamous Intra-epithelial Lesion              & 100            & 100            & 100            & 100            \\ \cline{2-6} 
                                       & Negative for Intra epithelial Malignancy          & 100            & 100            & 100            & 100            \\ \cline{2-6} 
                                       & Squamous Cell Carcinoma    & 88.24          & 100            & 93.75          & 100            \\ \cline{2-6} 
                                       & \textbf{Aggregrate}        & \textbf{99.08} & \textbf{98.95} & \textbf{98.97} & \textbf{99.48} \\ \hline
\multirow{6}{*}{\textbf{SIPaKMeD WSI}} & Dyskeratotic               & 91.67          & 97.78          & 94.62          & 97.78          \\ \cline{2-6} 
                                       & Koilocytotic               & 100            & 87.5           & 93.33          & 87.5           \\ \cline{2-6} 
                                       & Metaplastic                & 96.36          & 96.36          & 96.36          & 96.36          \\ \cline{2-6} 
                                       & Parabasal                  & 100            & 100            & 100            & 100            \\ \cline{2-6} 
                                       & Superficial   Intermediate & 89.66          & 100            & 94.55          & 100            \\ \cline{2-6} 
                                       & \textbf{Aggregrate}        & \textbf{95.54} & \textbf{96.33} & \textbf{95.77} & \textbf{96.33} \\ \hline
\multirow{6}{*}{\textbf{SIPaKMeD SCI}} & Dyskeratotic               & 98.18          & 99.39          & 98.78          & 99.39          \\ \cline{2-6} 
                                       & Koilocytotic               & 98.75          & 95.18          & 96.93          & 95.15          \\ \cline{2-6} 
                                       & Metaplastic                & 95.73          & 98.74          & 97.21          & 98.74          \\ \cline{2-6} 
                                       & Parabasal                  & 100            & 100            & 100            & 100            \\ \cline{2-6} 
                                       & Superficial   Intermediate & 100            & 99.4           & 99.7           & 99.4           \\ \cline{2-6} 
                                       & \textbf{Aggregrate}        & \textbf{98.55} & \textbf{98.53} & \textbf{98.53} & \textbf{98.54} \\ \hline
\end{tabular}
\end{table*}

\begin{table*}[]
\centering
\caption{Results obtained by the base classifiers (before fusion) on the three datasets used in this research.}
\label{res_before}
\begin{tabular}{|c|c|c|c|c|c|}
\hline
\textbf{Dataset} & \textbf{Model} & \textbf{Accuracy(\%)} & \textbf{Precision(\%)} & \textbf{Recall(\%)} & \textbf{F1-Score(\%)} \\ \hline
\multirow{3}{*}{Mendeley LBC} & Inception v3 & 97.96 & 97.95 & 97.56 & 97.75 \\ \cline{2-6} 
                              & DenseNet-161 & 98.44 & 97.06 & 98.44 & 97.63 \\ \cline{2-6} 
                              & ResNet-34    & 98.12 & 97.96 & 98.12 & 98.04 \\ \hline
\multirow{3}{*}{SIPaKMeD WSI} & Inception v3 & 88.27 & 88.63 & 89.97 & 89.13 \\ \cline{2-6} 
                              & DenseNet-161 & 93.08 & 92.38 & 93.08 & 92.69 \\ \cline{2-6} 
                              & ResNet-34    & 93.28 & 92.22 & 93.29 & 92.62 \\ \hline
\multirow{3}{*}{SIPaKMeD SCI} & Inception v3 & 94.34 & 94.31 & 94.38 & 94.31 \\ \cline{2-6} 
                              & DenseNet-161 & 97.28 & 97.28 & 97.29 & 97.28 \\ \cline{2-6} 
                              & ResNet-34    & 97.17 & 97.27 & 97.22 & 97.19 \\ \hline
\end{tabular}
\end{table*}

The datasets used in the present study have been split into 3:1:1 ratio of train, validation and test sets. The three pre-trained CNN models have been fine-tuned using the datasets by freezing the weights of the top 5 layers and training for 50 epochs. The probability distributions of the models have been saved and fused for the final classification using the Sugeno Fuzzy Integral. The confusion matrices thus obtained on the test sets of the respective datasets are shown in \autoref{cm}. Consequently, the class-wise results and the aggregate results of all the class are tabulated in \autoref{res_data} for the three datasets used. The results obtained before the ensemble, that is the results obtained by the base classifiers are shown in \autoref{res_before}.
\subsection{Verification of Complementarity of Features}
\begin{table}[]
\centering
\caption{KL and JS Divergences between individual models on Mendeley LBC Dataset}
\label{kl_mendeley}
\begin{tabular}{|c|c|c|c|}
\hline
\textbf{Distribution P} & \textbf{Distribution Q} & \textbf{D(P||Q)} & \textbf{D(P||Q)}       \\ \hline
Inception v3            & DenseNet-161             & 2.356            & \multirow{2}{*}{0.131} \\ \cline{1-3}
DenseNet-161             & Inception v3            & 0.611            &                        \\ \hline
Inception v3            & ResNet-34                & 6.055            & \multirow{2}{*}{0.115} \\ \cline{1-3}
ResNet-34                & Inception v3            & 0.540            &                        \\ \hline
DenseNet-161             & ResNet-34                & 3.300            & \multirow{2}{*}{0.162} \\ \cline{1-3}
ResNet-34                & DenseNet-161             & 0.884            &                        \\ \hline
\end{tabular}
\end{table}
To verify the complementary or dissimilar nature of the features of the pre-trained models used to extract the confidence scores of the datasets, two statistical divergence metrics are used: the Kullback-Leibler Divergence (KLD) \cite{kullback1951information, kullback1997information} and the Jensen-Shannon Divergence (JSD) \cite{menendez1997jensen}.

\begin{table}[]
\centering
\caption{KL and JS Divergences between individual models on SIPaKMeD WSI Dataset}
\label{kl_sipakslide}
\begin{tabular}{|c|c|c|c|}
\hline
\textbf{Distribution P} & \textbf{Distribution Q} & \textbf{D(P||Q)} & \textbf{D(P||Q)}       \\ \hline
Inception v3            & DenseNet-161             & 0.502            & \multirow{2}{*}{0.152} \\ \cline{1-3}
DenseNet-161             & Inception v3            & 0.367            &                        \\ \hline
Inception v3            & ResNet-34                & 0.584            & \multirow{2}{*}{0.156} \\ \cline{1-3}
ResNet-34                & Inception v3            & 0.418            &                        \\ \hline
DenseNet-161             & ResNet-34                & 0.212            & \multirow{2}{*}{0.132} \\ \cline{1-3}
ResNet-34                & DenseNet-161             & 0.208            &                        \\ \hline
\end{tabular}
\end{table}

The KLD is a non-symmetric measure of dissimilarity between two probability distributions on the same probability space. Let there be a probability space $S$, and two probability distributions on this space $p(s)>0$ and $q(s)>0$ for every discrete variable $s \in S$, such that $p(s)+q(s)=1$. Then the discrete form of KLD defined from $q(s)$ to $p(s)$ is given as \autoref{kld}, $\log(x)$ being the natural logarithm of $x$.
\begin{equation}\label{kld}
     D\textsubscript{KL}\left(p(s)||q(s) \right) = \sum_{s\in S}p(s).\log \left(\frac{p(s)}{q(s)}\right)
\end{equation}

\begin{table}[]
\centering
\caption{KL and JS Divergences between individual models on SIPaKMeD SCI Dataset}
\label{kl_sipakcell}
\begin{tabular}{|c|c|c|c|}
\hline
\textbf{Distribution P} & \textbf{Distribution Q} & \textbf{D(P||Q)} & \textbf{D(P||Q)}       \\ \hline
Inception v3            & DenseNet-161             & 0.355            & \multirow{2}{*}{0.129} \\ \cline{1-3}
DenseNet-161             & Inception v3            & 0.160            &                        \\ \hline
Inception v3            & ResNet-34                & 0.250            & \multirow{2}{*}{0.134} \\ \cline{1-3}
ResNet-34                & Inception v3            & 0.337            &                        \\ \hline
DenseNet-161             & ResNet-34                & 0.178            & \multirow{2}{*}{0.118} \\ \cline{1-3}
ResNet-34                & DenseNet-161             & 0.211            &                        \\ \hline
\end{tabular}
\end{table}

As $D\textsubscript{KL}\left(p(s)||q(s) \right) \neq D\textsubscript{KL}\left(q(s)||p(s) \right)$, a symmetrical statistical divergence have been derived from the KLD, called the Jensen-Shannon Divergence (JSD). JSD is effectively, a smoothed form of KLD. For the same probability distributions $p(s)$ and $q(s)$ as mentioned above, let $m(s)$ be another probability distribution such that $m(s) = \frac{p(s)+q(s)}{2}$. Then the JSD (discrete form) is given by \autoref{jsd}.
\begin{equation}\label{jsd}
    D\textsubscript{JS}(p(s)||q(s)) = \frac{\left[D\textsubscript{KL}(p(s)||m(s))+D\textsubscript{KL}(q(s)||m(s))\right]}{2}
\end{equation}
The KLD and JSD measures between the decision scores of each pair of CNN classifiers, are shown in Tables \ref{kl_mendeley}, \ref{kl_sipakslide} and \ref{kl_sipakcell} for the Mendeley, SIPaKMeD WSI and SIPaKMeD SCI datasets respectively.
\subsection{Comparison with different back-bone CNNs}
\begin{figure*}
    \centering
    \includegraphics[scale=0.5]{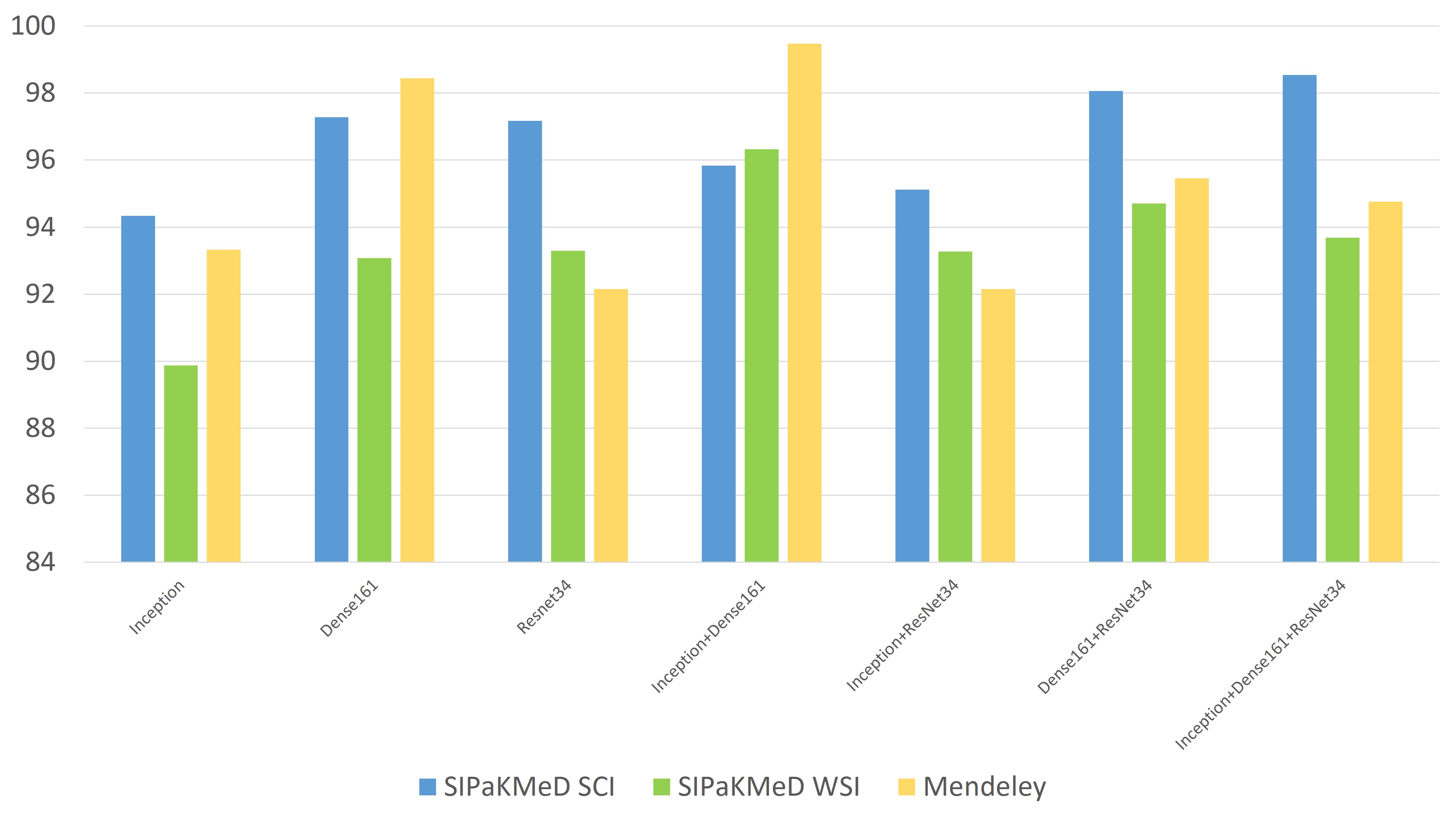}
    \caption{Comparison with different CNN architectures}
    \label{backbones}
\end{figure*}

It has been mentioned earlier that we have evaluated fuzzy measure over three popularly used pre-trained CNNs such that Inception v3, DenseNet-161 and ResNet-34. The results obtained by these datasets on all three datasets is given by \autoref{backbones}.  It can be observed that for SIPaKMeD SCI the maximum is achieved by combining the probability distribution of all three neural nets. Whereas for SIPaKMeD WSI and Mendeley LBC dataset ensemble of Inception v3 and DenseNet-161 datasets achieve the most.

\subsection{Comparison with Other Ensemble Approaches}
\begin{table}[]
\centering
\caption{Comparison of accuracies obtained by Fuzzy Fusion with other popular ensemble techniques}
\label{comp_ensemble}
\resizebox{0.6\columnwidth}{!}{
\begin{tabular}{|c|c|c|c|}
\hline
\textbf{\begin{tabular}[c]{@{}c@{}}Ensemble \\ Technique\end{tabular}} &
  \textbf{\begin{tabular}[c]{@{}c@{}}Mendeley \\ LBC\\ Dataset(\%)\end{tabular}} &
  \textbf{\begin{tabular}[c]{@{}c@{}}SIPaKMeD \\ WSI\\ Dataset(\%)\end{tabular}} &
  \textbf{\begin{tabular}[c]{@{}c@{}}SIPaKMeD \\ SCI\\ Dataset(\%)\end{tabular}} \\ \hline
Majority Voting                                                          & 95.68          & 94.37          & 97.64          \\ \hline
Average                                                                  & 94.76          & 93.88          & 97.29          \\ \hline
\begin{tabular}[c]{@{}c@{}}Weighted \\ Average\end{tabular}              & 98.96          & 95.11          & 98.03          \\ \hline
Product Rule                                                             & 92.15          & 93.37          & 97.29          \\ \hline
Maximum Rule                                                             & 92.15          & 94.89          & 97.54          \\ \hline
\begin{tabular}[c]{@{}c@{}}Choquet Fuzzy\\ Integral\end{tabular}         & 98.96          & 95.41          & 98.40          \\ \hline
\textbf{\begin{tabular}[c]{@{}c@{}}Sugeno Fuzzy\\ Integral\end{tabular}} & \textbf{99.48} & \textbf{96.33} & \textbf{98.54} \\ \hline
\end{tabular}
}
\end{table}
The same probability distributions have been used to compute the predictions on the datasets using some popular ensembling procedures. The results thus obtained have been tabulated in \autoref{comp_ensemble}. Among the ensemble techniques used, the weighted probability average ensemble with weights \{0.5, 2.0, 1.0\} for \{Inception v3, DenseNet-161, ResNet-34\} (weights set experimentally), gave results closest to the Sugeno Fuzzy Integral ensemble. The fuzzy measures used for the Sugeno Fuzzy Integral are \{Inception v3, DenseNet-161\}$=\{1, 0.5\}$ for Mendeley LBC and SIPaKMeD WSI datasets and \{Inception v3, DenseNet-161, ResNet-34\}$=\{0.5, 0.5, 0.1\}$ for SIPaKMeD SCI dataset. The fuzzy measures have been set through extensive experiments on multiple runs of the framework.
\subsection{Results with different fuzzy measures}
\begin{table}[]
\centering
\caption{Results obtained by different fuzzy measures of the fuzzy integral ensemble on SIPaKMeD SCI dataset}
\begin{tabular}{|c|c|c|c|}
\hline
\multicolumn{3}{|c|}{\textbf{Fuzzy Measures}}                                                                 & \multicolumn{1}{c|}{\multirow{2}{*}{\textbf{Accuracy}}} \\ \cline{1-3}
\multicolumn{1}{|c|}{\textbf{Inception v3}} & \multicolumn{1}{c|}{\textbf{DenseNet-161}} & \multicolumn{1}{c|}{\textbf{ResNet-34}} & \multicolumn{1}{c|}{}                          \\ \hline
0.5                               & 0.5                              & 0.1                           & 95.33                                          \\ \hline
0.5                               & 0.1                              & 0.5                           & 95.36                                          \\ \hline
\textbf{0.1}                      & \textbf{0.5}                     & \textbf{0.5}                  & \textbf{98.54}                                 \\ \hline
1                                 & 0.5                              & 0.5                           & 97.52                                          \\ \hline
0.5                               & 1                                & 0.5                           & 95.36                                          \\ \hline
0.5                               & 0.5                              & 1                             & 97.57                                          \\ \hline
0.5                               & 1                                & 0.1                           & 97.52                                          \\ \hline
\end{tabular}
\label{fuzzy_sci}
\end{table}

\begin{table}[]
\centering
\caption{Results obtained by different fuzzy measures of the fuzzy integral ensemble on SIPaKMeD WSI dataset}
\begin{tabular}{|c|c|c|}
\hline
\multicolumn{2}{|c|}{\textbf{Fuzzy Measures}}                            & \multicolumn{1}{c|}{\multirow{2}{*}{\textbf{Accuracy}}} \\ \cline{1-2}
\multicolumn{1}{|c|}{\textbf{Inception v3}} & \multicolumn{1}{c|}{\textbf{DenseNet-161}} & \multicolumn{1}{c|}{}                          \\ \hline
\textbf{1}                      & \textbf{0.5}                  & \textbf{96.33}                                 \\ \hline
0.5                             & 1                             & 90.31                                          \\ \hline
0.5                             & 0.1                           & 30.31                                          \\ \hline
0.1                             & 0.5                           & 94.36                                          \\ \hline
\end{tabular}
\label{fuzzy_wsi}
\end{table}

\begin{table}[]
\centering
\caption{Results obtained by different fuzzy measures of the fuzzy integral ensemble on Mendeley LBC dataset}
\begin{tabular}{|c|c|c|}
\hline
\multicolumn{2}{|c|}{\textbf{Fuzzy Measures}}                            & \multicolumn{1}{c|}{\multirow{2}{*}{\textbf{Accuracy}}} \\ \cline{1-2}
\multicolumn{1}{|c|}{\textbf{Inception v3}} & \multicolumn{1}{c|}{\textbf{DenseNet-161}} &                               \\ \hline
\textbf{1}                      & \textbf{0.5}                  & \textbf{99.48}                \\ \hline
0.5                             & 1                             & 79.48                         \\ \hline
0.5                             & 0.1                           & 79.58                         \\ \hline
0.1                             & 0.5                           & 87.96                         \\ \hline
\end{tabular}
\label{fuzzy_mendeley}
\end{table}
Different experiments with the fuzzy measures have been conducted and the best set of weights have been chosen. The variation of accuracies with the change in fuzzy measures are shown in Tables \ref{fuzzy_sci}, \ref{fuzzy_wsi} and \ref{fuzzy_mendeley} for the SIPaKMeD SCI, SIPaKMeD WSI and the Mendeley LBC datasets respectively.
\subsection{Comparison with Existing Models}
\begin{table*}[]
\centering
\caption{Comparison of the proposed framework with existing methods in literature on the SIPaKMeD SCI and WSI datasets}
\label{litcomp}
\begin{tabular}{|c|c|c|c|}
\hline
\multirow{2}{*}{\textbf{Work}}         & \multirow{2}{*}{\textbf{Approach}}        & \multicolumn{2}{c|}{\textbf{Accuracy   (\%)}} \\ \cline{3-4} 
                                       &                                           & \textbf{SCI}          & \textbf{WSI}          \\ \hline
Kiran GV et al. \cite{gv2019automatic} & Feature   Extraction and PCA              & \textbf{99.63}        & \textbf{96.37}        \\ \hline
Shi et al. \cite{shi2019graph}         & Graph   Convolutional Network             & 98.37                 & -                     \\ \hline
Plissiti et al. \cite{plissiti2018sipakmed} & \begin{tabular}[c]{@{}c@{}}Features:   Deep fully CNNs\\      Classifiers: SVM  and CNNs\end{tabular}       & 95.35 & - \\ \hline
Win et al. \cite{win2020computer}           & \begin{tabular}[c]{@{}c@{}}Features:   RF\\      Classifiers: LDA, SVM, KNN and Decision trees\end{tabular} & 94.09 & - \\ \hline
Sevi et al. \cite{sevihealth}          & CNNs                                      & 88.40                 & -                     \\ \hline
\textbf{Proposed approach}             & \textbf{Sugeno   Fuzzy Integral Ensemble} & 98.54        & 96.33        \\ \hline
\end{tabular}
\end{table*}

Here in this section, we have given the comparative study of performances of our proposed approach with previously reported works. In the Mendeley dataset, no works have been reported so far, therefore the works on the SIPaKMeD dataset are presented for comparison purpose. In \cite{kiran2019automatic}, the reported accuracy in SIPaKMeD WSI is 96.37\% which is almost the same as ours. The comparative results in SIPaKMeD are given by \ref{litcomp}. Shi et al \cite{shi2019graph} achieves impressive result of 98.37\% on 5-class SIPaKMeD dataset. Kiran GV et al. \cite{gv2019automatic} extracted features from the ResNet-34 CNN model using transfer learning and applied Principal Component Analysis in the penultimate feature layer of the CNN for the final feature set selection and classification. They achieved an accuracy of 99.63\% on the SIPaKMeD SCI dataset and 96.37\% on the SIPaKMeD WSI dataset employing 5-fold cross-validation. However, we have implemented a different approach that does not require extraction of features and takes the opinion of multiple experts (CNN models) making the performance robust for the different datasets used, and is computationally efficient while keeping classification performance at par with state-of-the-art. It is seen that the proposed approach outperforms most of the works evolved so far which justifies the reliability of the model.

\subsection{GradCAM Analysis}

\begin{figure*}
    \centering
    \subfloat[Mendeley: Original]{\includegraphics[scale = 0.07]{pics/mendeley.jpg}}\;
    \subfloat[Mendeley: Inception v3]{\includegraphics[scale = 0.052]{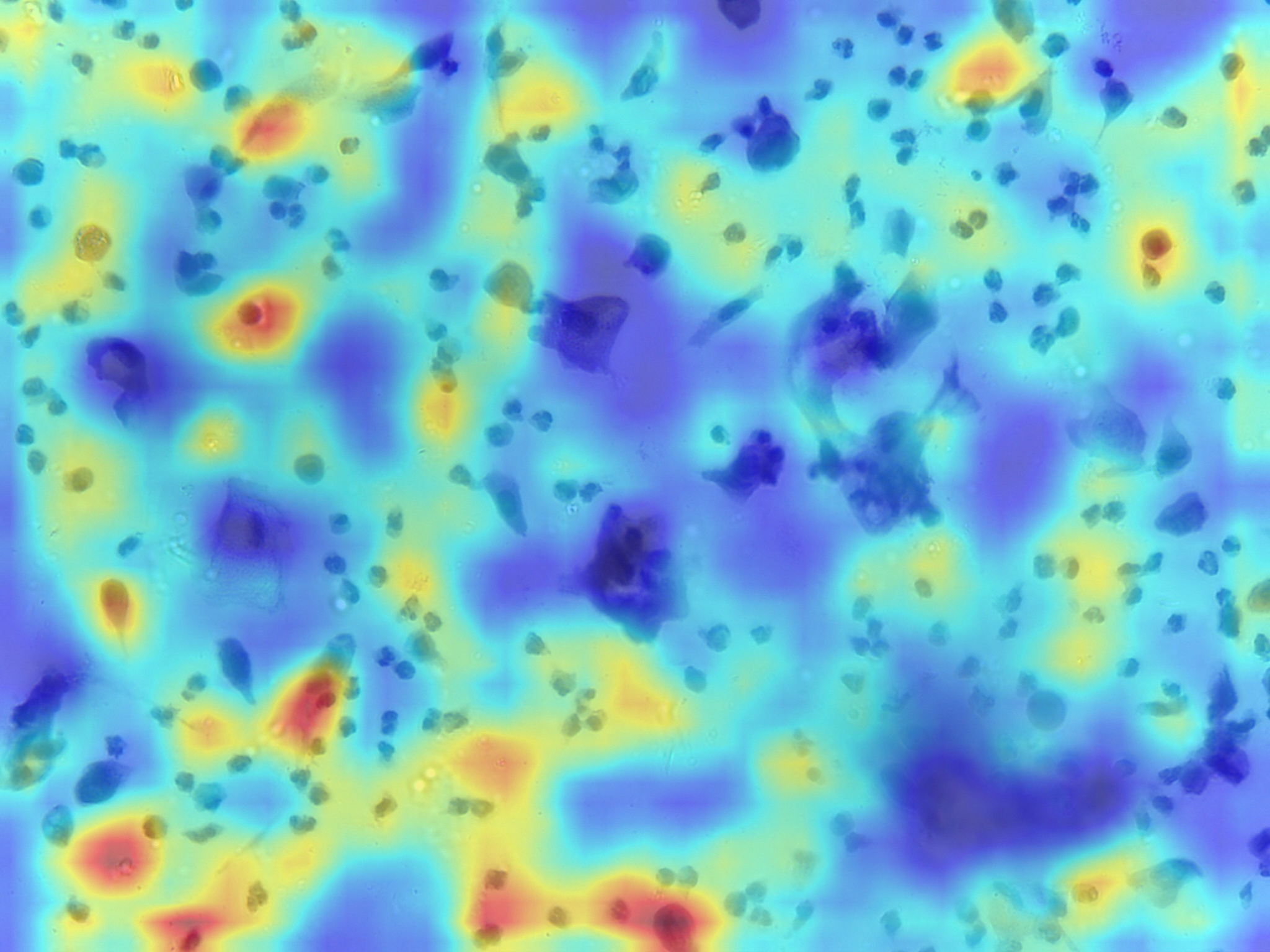}}\;
    \subfloat[Mendeley: DenseNet-161]{\includegraphics[scale = 0.052]{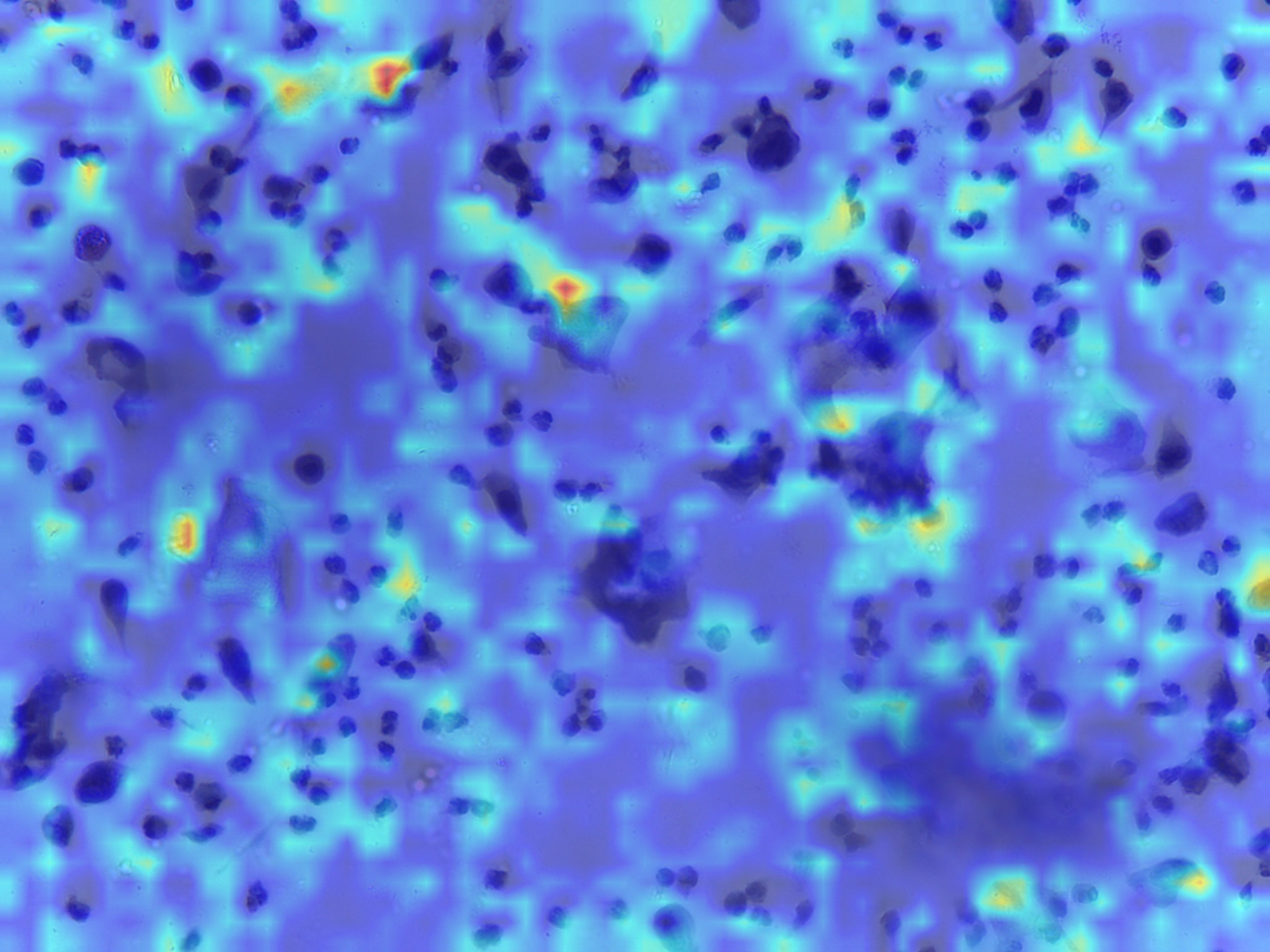}}\;
    \subfloat[Mendeley: ResNet-34]{\includegraphics[scale=0.052]{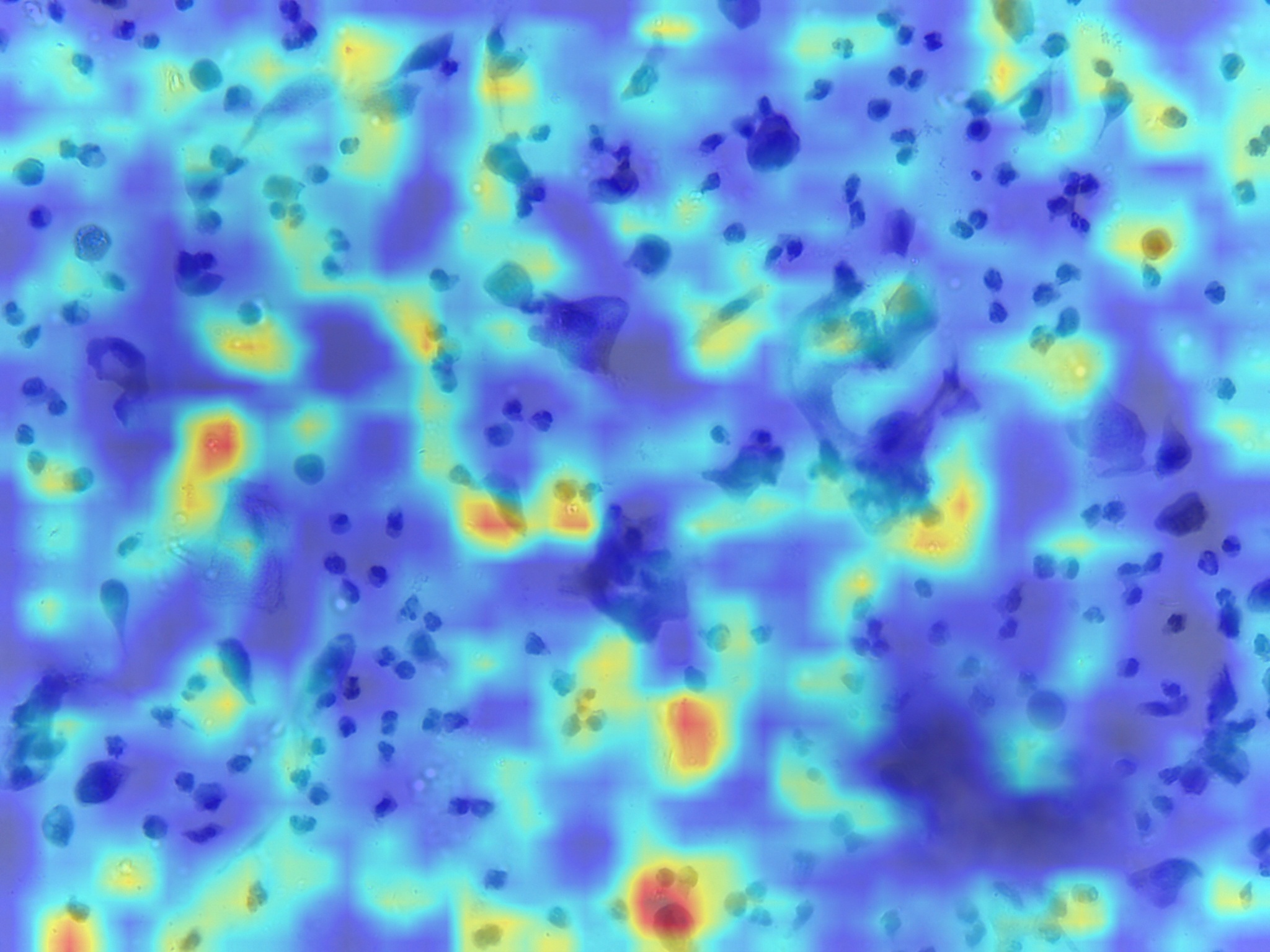}}\\
    
    \subfloat[SIPaKMeD WSI: Original]{\includegraphics[scale = 0.055]{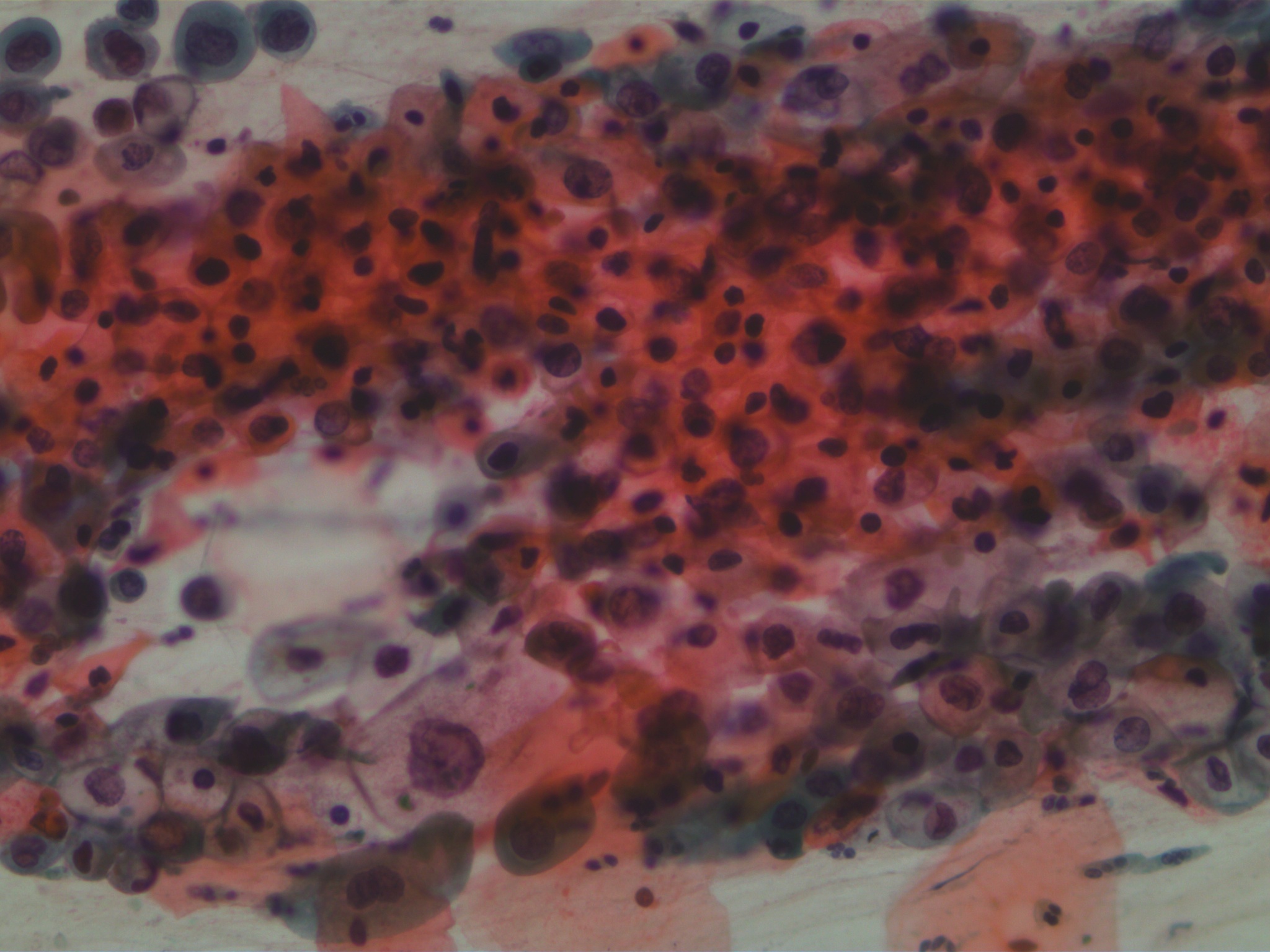}}\;
    \subfloat[SIPaKMeD WSI: Inception v3]{\includegraphics[scale = 0.055]{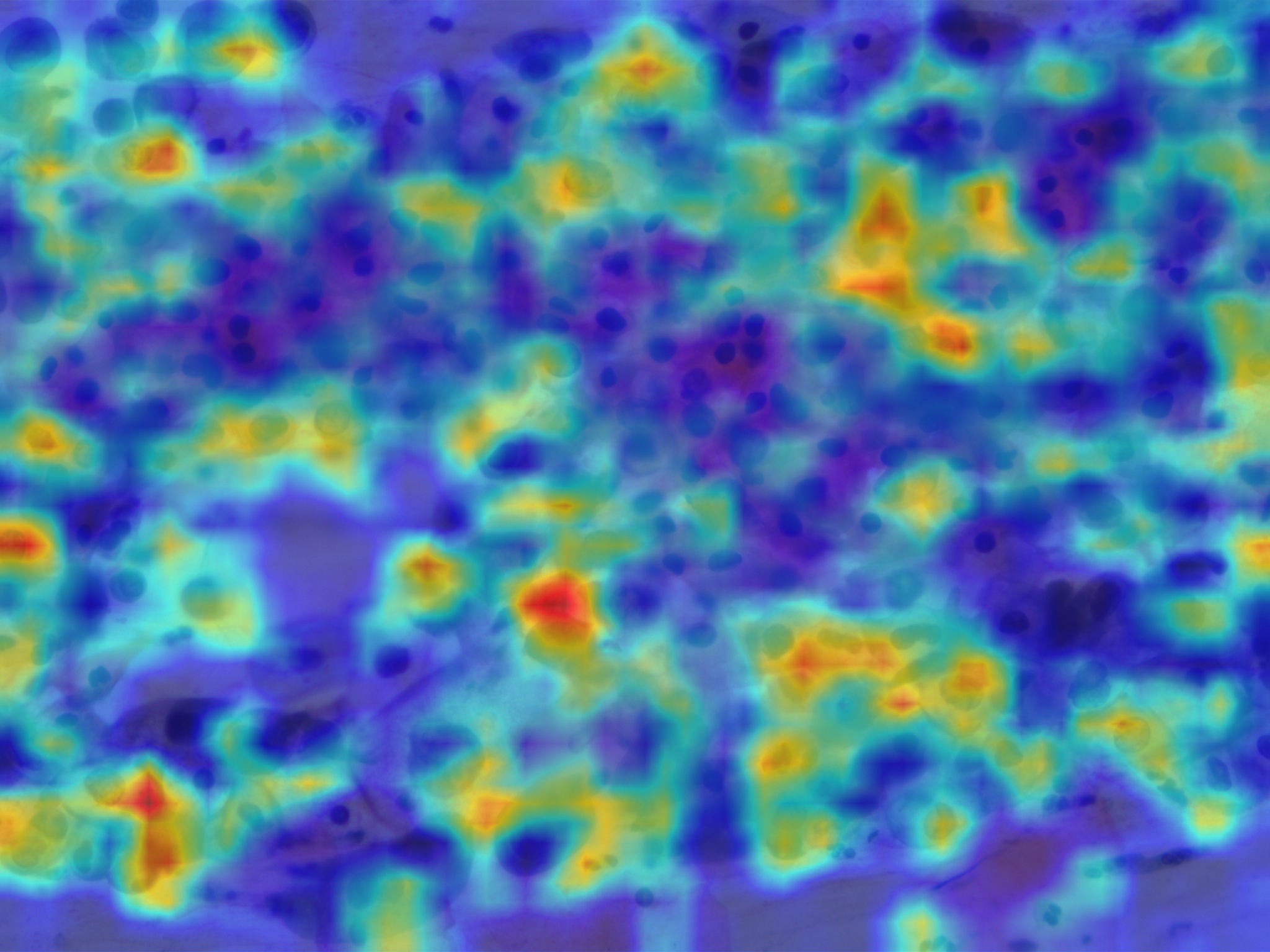}}\;
    \subfloat[SIPaKMeD WSI: DenseNet-161]{\includegraphics[scale = 0.055]{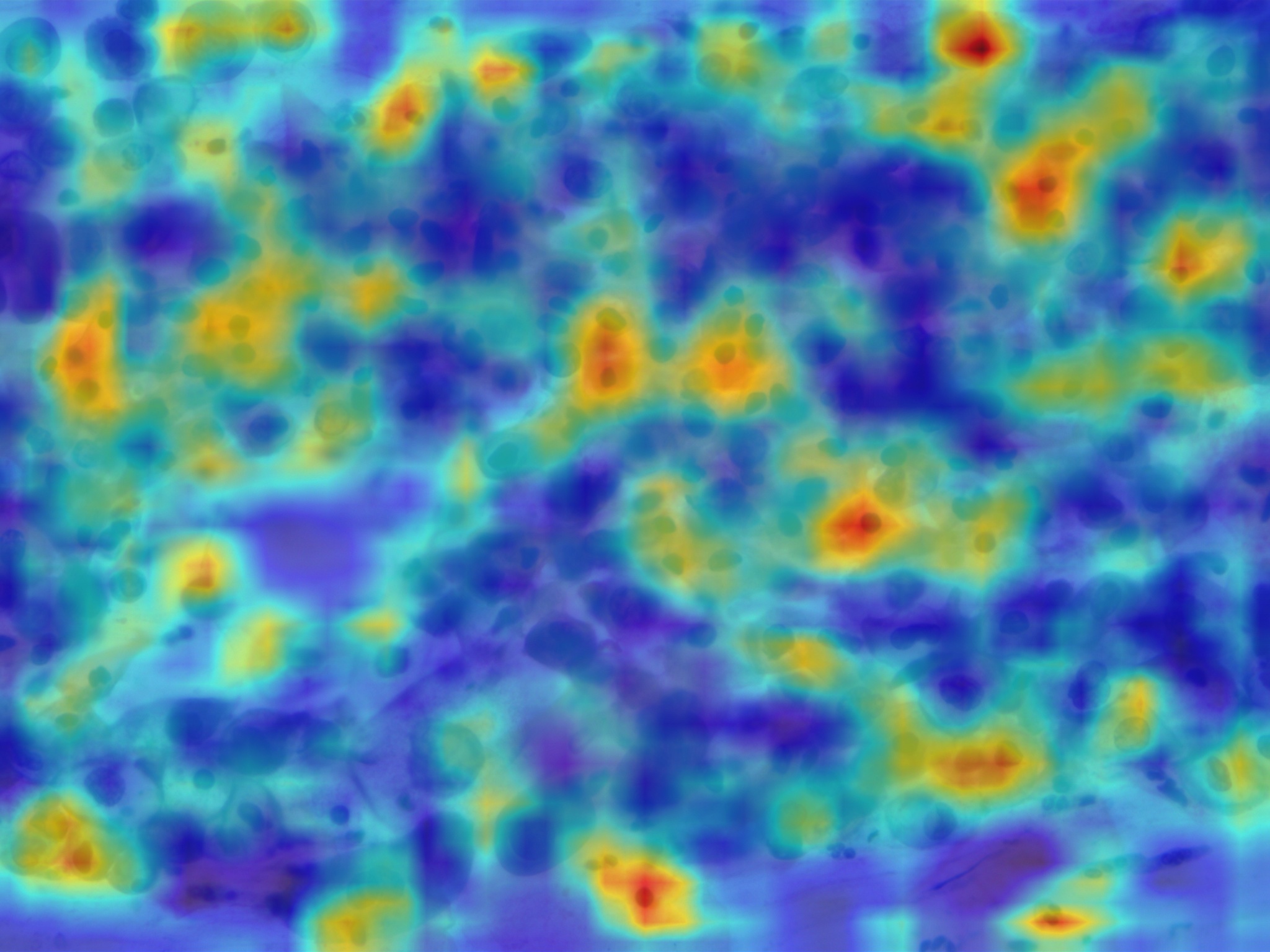}}\;
    \subfloat[SIPaKMeD WSI: ResNet-34]{\includegraphics[scale = 0.055]{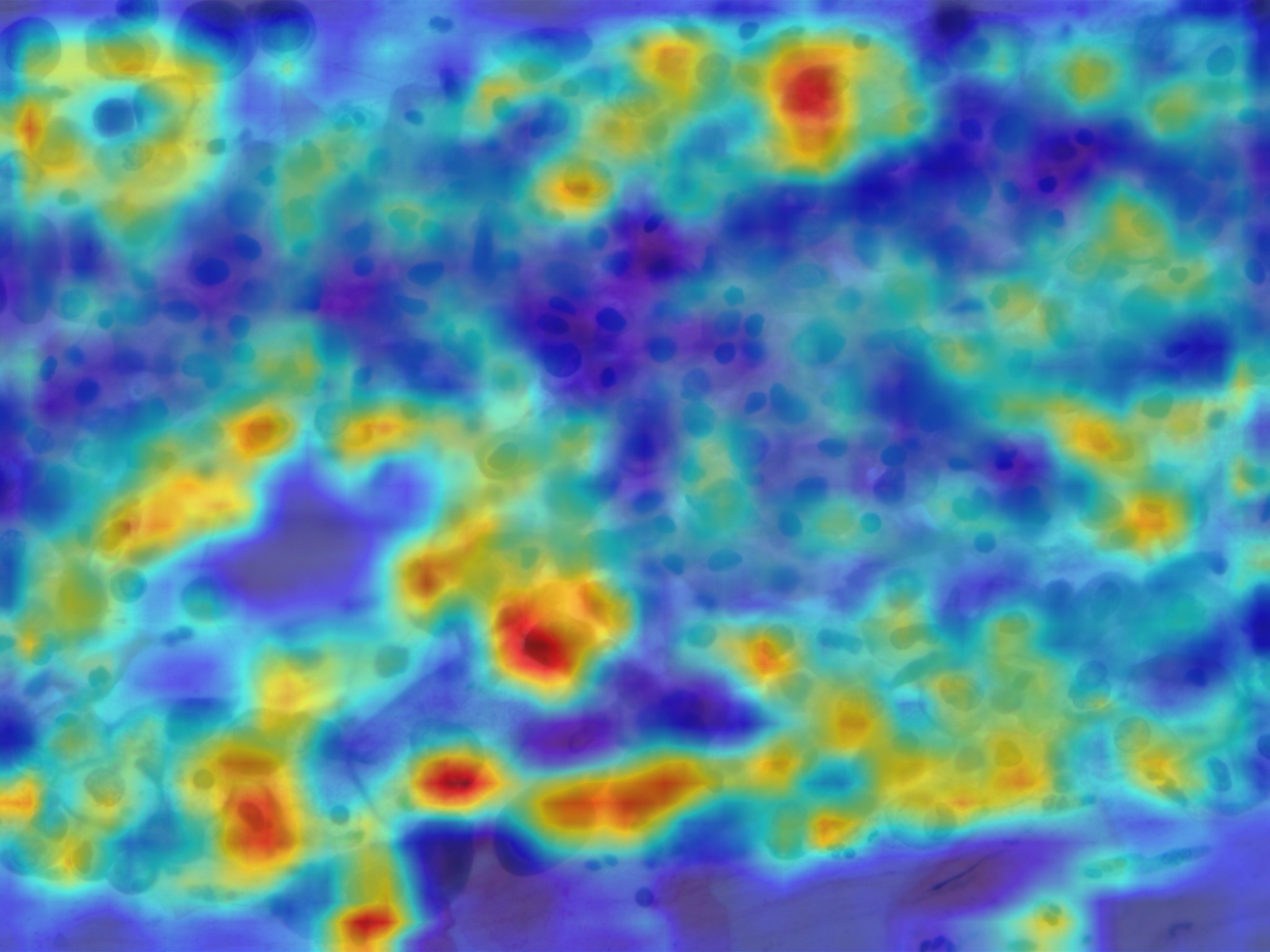}}\\
    
    \subfloat[SIPaKMeD SCI: Original]{\includegraphics[scale = 0.46]{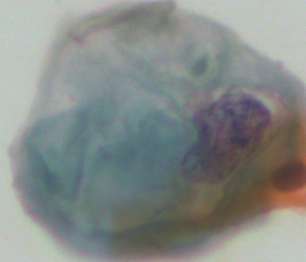}}\;
    \subfloat[SIPaKMeD SCI: Inception v3]{\includegraphics[scale = 0.35]{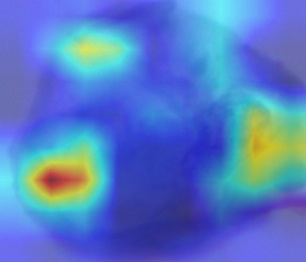}}\;
    \subfloat[SIPaKMeD SCI: DenseNet-161]{\includegraphics[scale = 0.35]{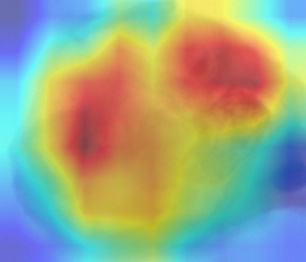}}\;
    \subfloat[SIPaKMeD SCI: ResNet-34]{\includegraphics[scale = 0.35]{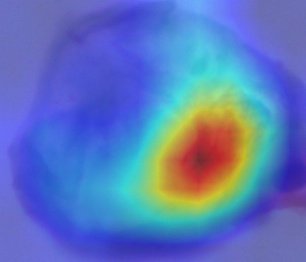}}
    
    \caption{GradCAM activations using the three base learners- Inception v3, DenseNet-161 and ResNet-34 on the three datasets used: (a)-(d) Mendeley LBC dataset, (e)-(h) SIPaKMeD WSI dataset and (i)-(l) SIPaKMeD SCI dataset.}
    \label{fig:gradcam}
\end{figure*}

In this section we use the Gradient guided Class Activation Maps or GradCAM by Selvaraju et al. \cite{selvaraju2017grad} to visually represent the distinguishing regions in the single cell and whole slide pap stained images that enables to make the base classifiers to make the predictions. The results for the same are show in \autoref{fig:gradcam} for the three datasets used in this study. GradCAM computes the number of weights in feature map of the last convolution layer to calculate the contribution of the feature maps towards the class prediction made by the CNN classifier. 

For all the three datasets, in \autoref{fig:gradcam}, it can be noted that the three classifiers focus on different regions of the corresponding original image. For example, in \autoref{fig:gradcam}(i), for the SIPaKMeD SCI dataset, the ResNet-34 model (\autoref{fig:gradcam}(l)) puts attention solely on the nucleus of the cell, DenseNet-161 (\autoref{fig:gradcam}(k)) focuses on the nucleus as well as on the cytoplasm of the cell. Inception v3 in \autoref{fig:gradcam}(j) focuses on the outliers and on the nucleus. Clearly, the three models takes into account different aspects of the image. Thus, when the ensemble of these three models are computer, the prediction incurs the complementary information provided by these different classifiers, and a superior prediction is made. Similarly for the whole slide images of Mendeley LBC dataset (\autoref{fig:gradcam}(a)) and SIPaKMeD WSI dataset (\autoref{fig:gradcam}(e)), different classifiers focus on different cells within the slide to compute the final predictions, which further enables the ensemble model to aggregate the discerning information from the base learners to compute a prediction.

\subsection{Error Analysis}
\begin{figure*}
    \centering
    \subfloat[SIPaKMeD SCI: Superficiel Intermediate]{\includegraphics[scale=0.5]{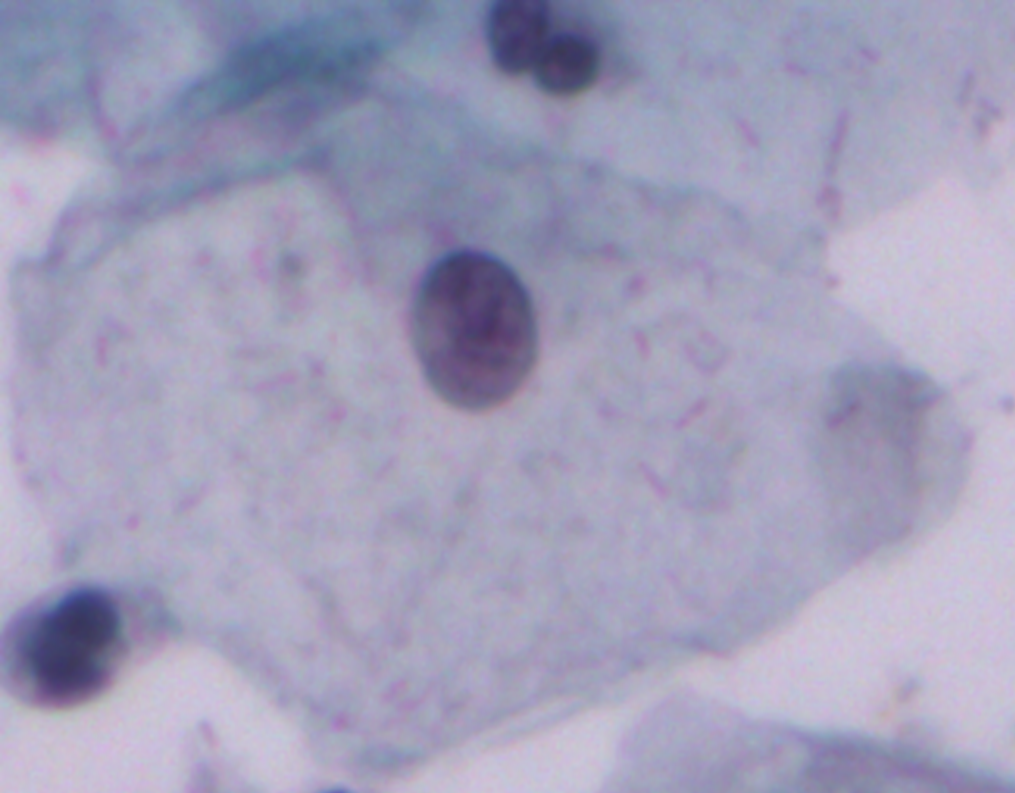}}\;\;
    \subfloat[SIPaKMeD WSI: Metaplastic]{\includegraphics[scale=0.5]{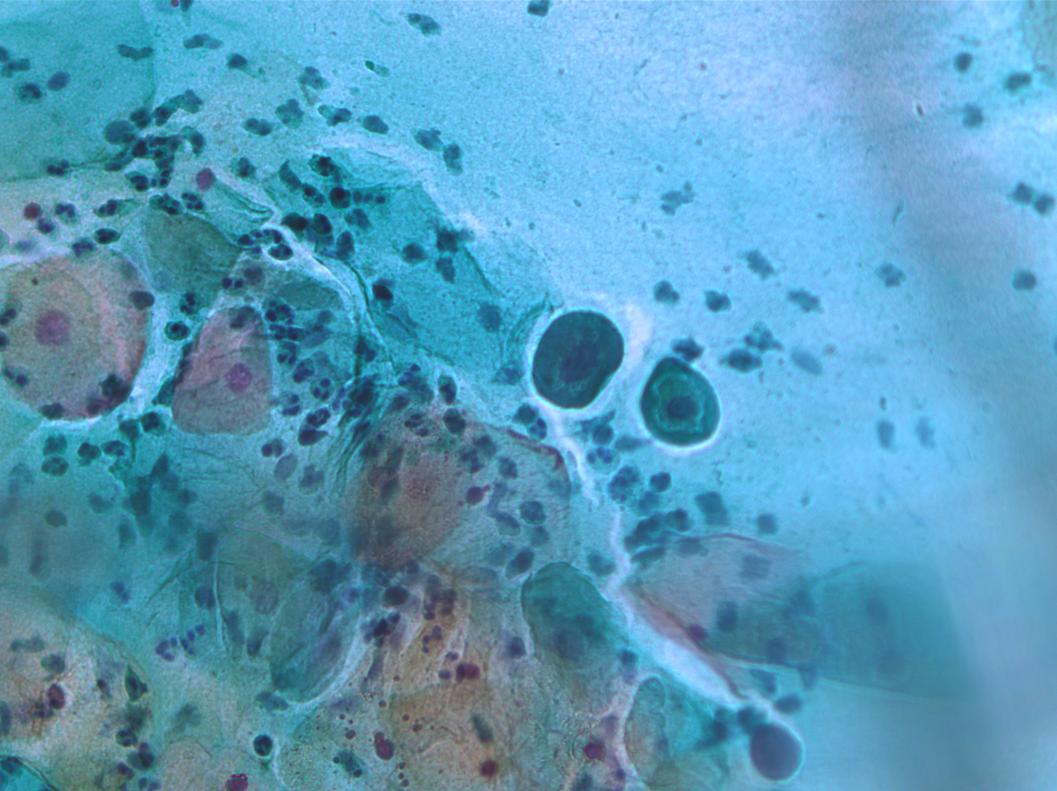}}\;\;
    \subfloat[Mendeley LBC: NILM]{\includegraphics[scale=0.5]{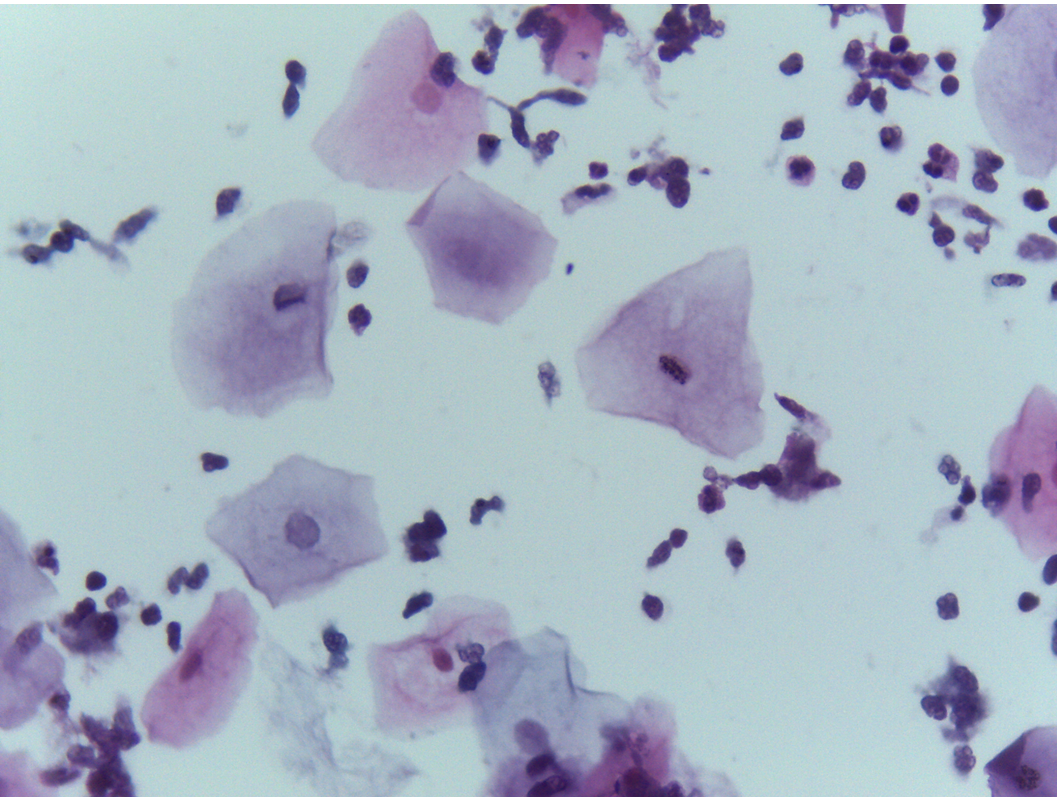}}
    \caption{Examples of instances where the ensemble approach made correct predictions although all contributing classifiers did not predict correctly}
    \label{correct}
\end{figure*}

The proposed framework shows robust and reliable performance in the cervical cytology image classification task. For example, \autoref{correct}, shows examples of instances where some individual classifiers predicted wrong classes while the ensemble approach made correct predictions. \autoref{correct}(a) shows a test image from the SIPaKMeD SCI dataset belonging to the class "Superficiel Intermediate", and predicted correctly by the fuzzy ensemble method despite the image containing multiple nuclei. But, this image was classified incorrectly by Inception v3 and ResNet-34 (but correctly by DenseNet-161). The confidence of DenseNet-161 on its prediction of this instance was much higher than Inception v3 and ResNet-34 on their predictions. This resulted in the ensemble method to give priority to the DenseNet-161 model's decision and predicting the sample to be "Superficiel Intermediate". Similarly, \autoref{correct}(b) shows a sample from the SIPaKMeD WSI dataset which was predicted correctly by DenseNet-161 to be "Metaplastic" but wrongly by Inception v3 as "Dyskeratotic". \autoref{correct}(c) shows a correct prediction from the Mendeley LBC dataset as belonging to the class "NILM", where DenseNet-161 made the correct prediction while Inception v3 predicted it to be "HSIL".

\begin{figure}
    \centering
    \includegraphics[scale=0.1]{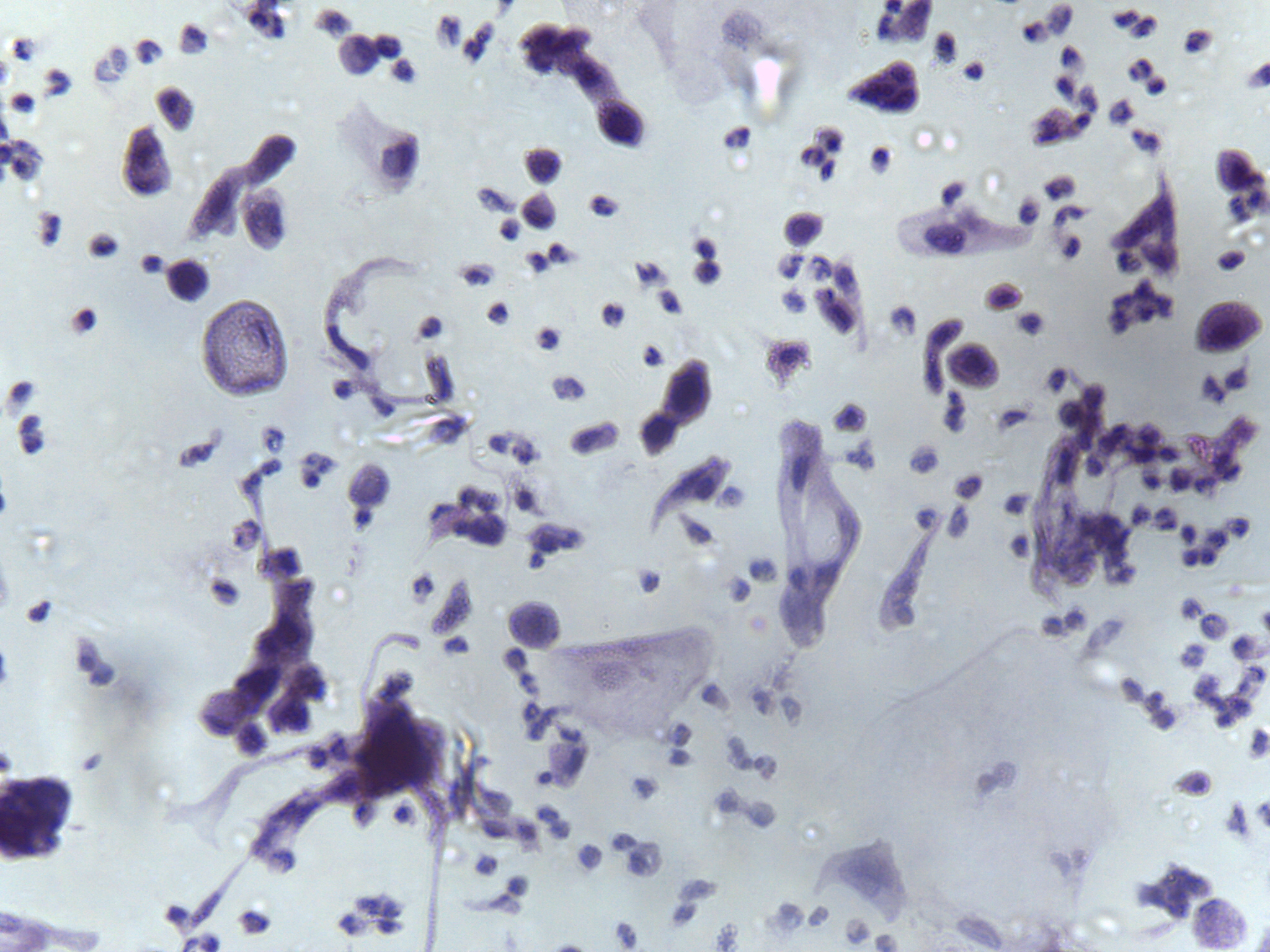}
    \caption{Misclassified HSIL class image of Mendeley LBC dataset}
    \label{mis_mendeley}
\end{figure}
\autoref{mis_mendeley} shows the only misclassified sample from the Mendeley LBC dataset. The image belongs to the class HSIL but was predicted to be of class SCC by the proposed model.

\begin{figure*}
    \centering
    \subfloat[Dyskeratotic]{\includegraphics[scale=0.15]{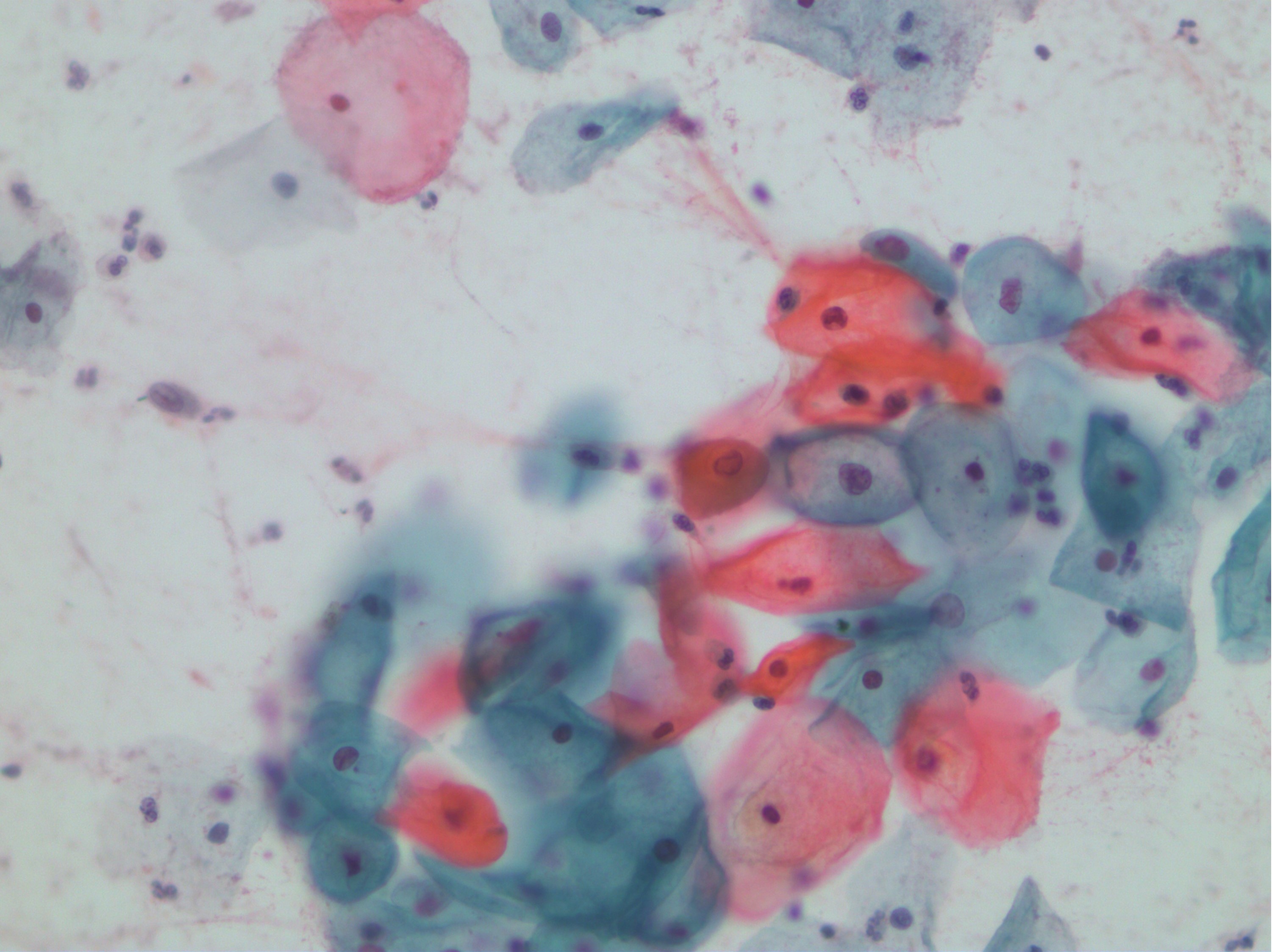}}\;
    \subfloat[Koilocytotic]{\includegraphics[scale=0.15]{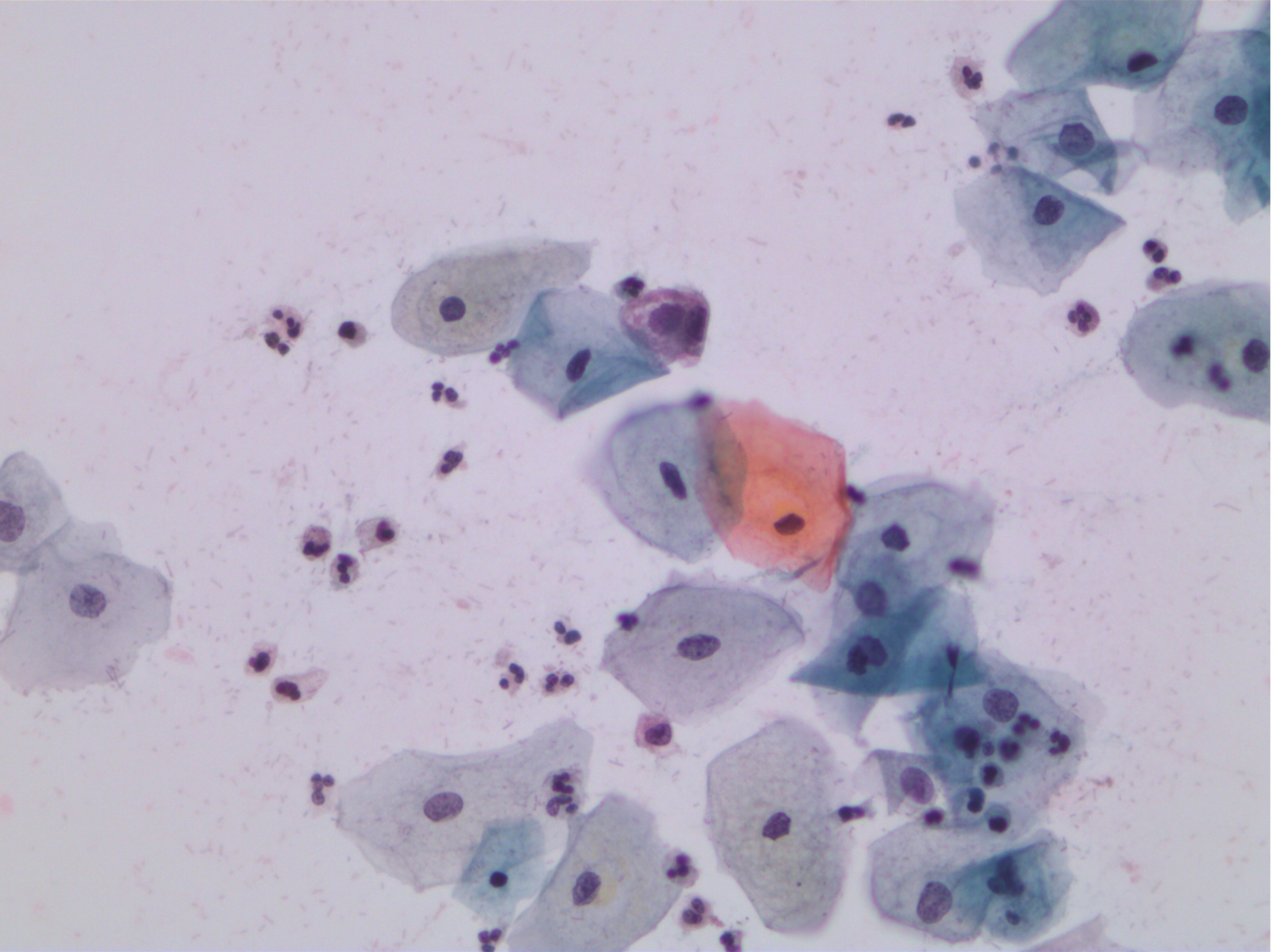}}\;
    \subfloat[Metaplastic]{\includegraphics[scale=0.15]{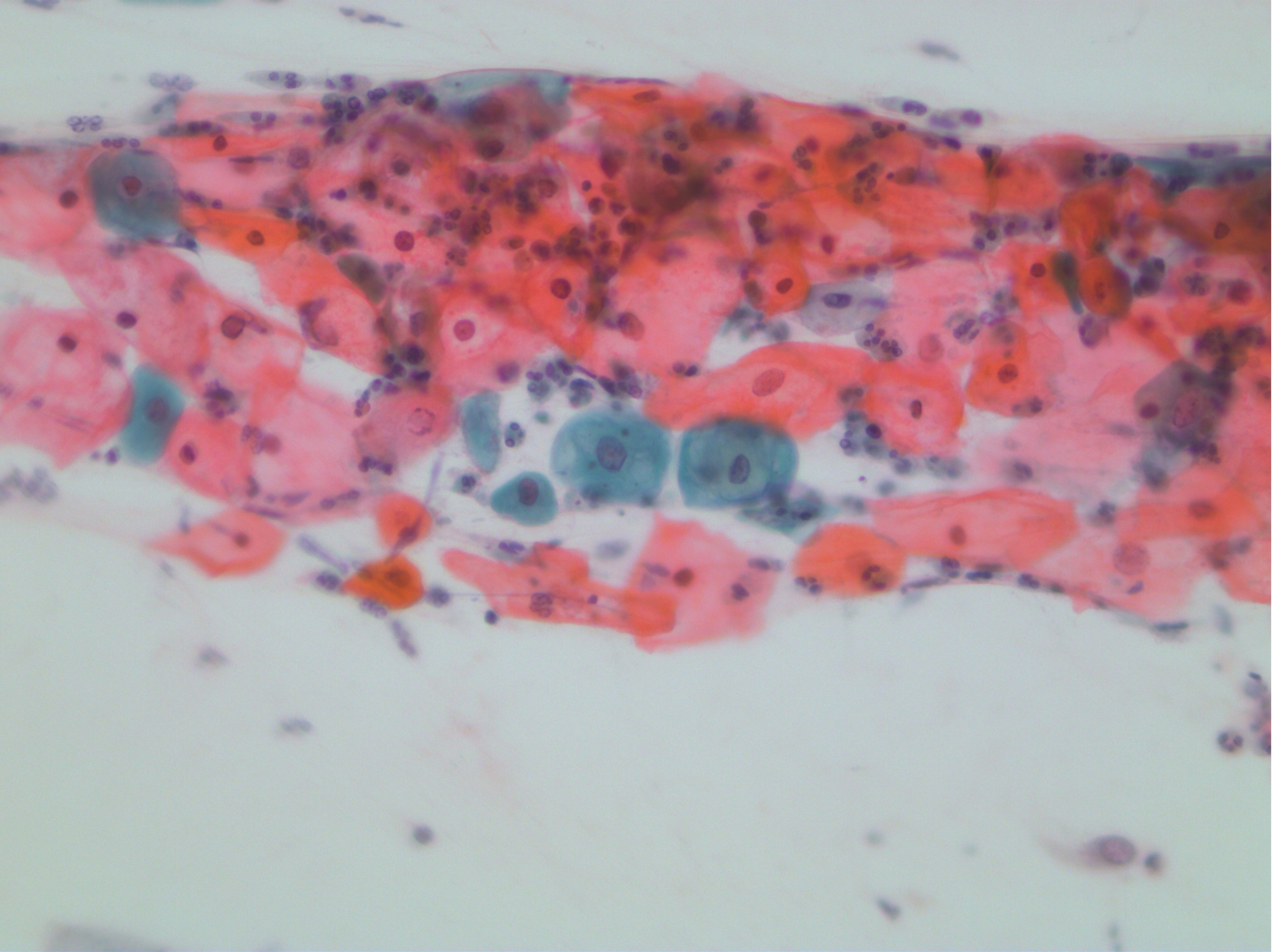}}
    \caption{Misclassified samples of SIPaKMeD WSI dataset, originally belonging to (a) Dyskeratotic (b) Koilocytotic and (c) Metaplastic classes}
    \label{mis_sipakslide}
\end{figure*}

\begin{figure*}
    \centering
    \subfloat[Dyskeratotic]{\includegraphics[scale=0.5]{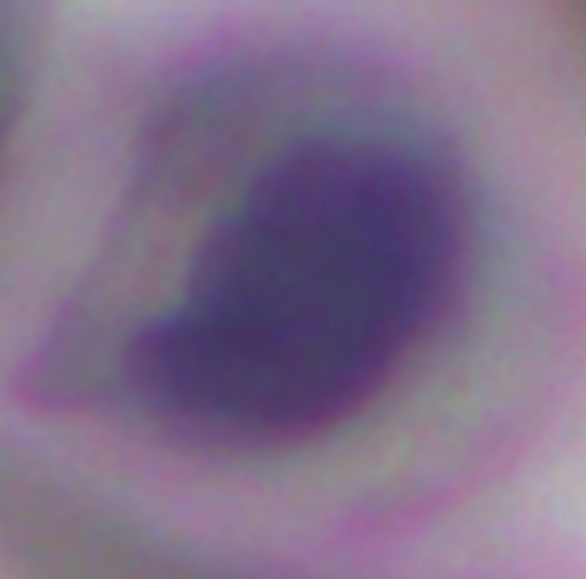}}\;
    \subfloat[Koilocytotic]{\includegraphics[scale=0.5]{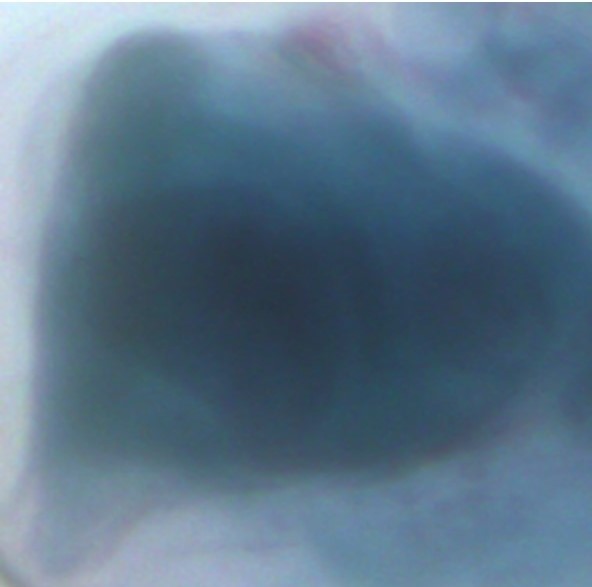}}\;
    \subfloat[Metaplastic]{\includegraphics[scale=0.5]{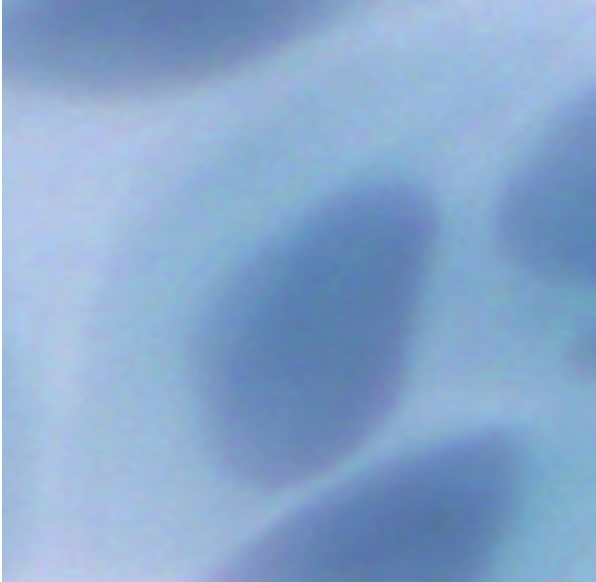}}\;
    \subfloat[Superficial Intermediate]{\includegraphics[scale=0.5]{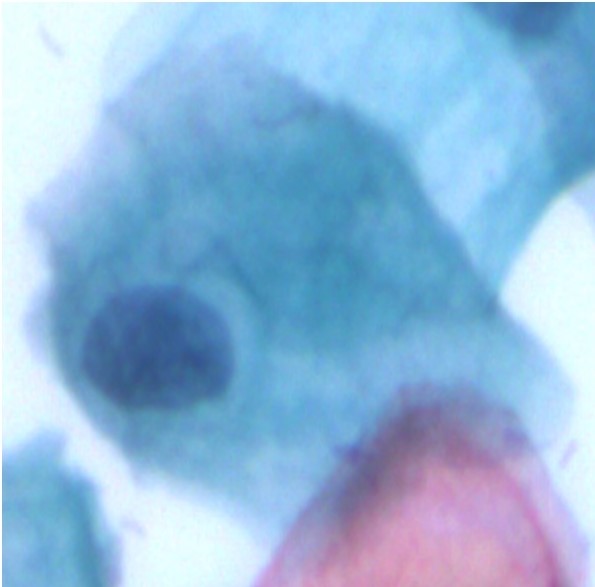}}
    \caption{Misclassified samples of SIPaKMeD SCI dataset, originally belonging to (a) Dyskeratotic (b) Koilocytotic (c) Metaplastic and (d) Superficial Intermediate classes}
    \label{mis_sipakcell}
\end{figure*}

\autoref{mis_sipakslide} shows some misclassified samples from the SIPaKMeD WSI dataset. The most probable reason for the wrong classification, in this case, is the presence of several types of cells in a single image. For example in \autoref{mis_sipakslide}(a), the number of "Metaplastic" cells is more than the number of "Dyskeratotic" cells, which led to the originally "Dyskeratotic" class image to be classified as "Metaplastic". The case is reversed for \autoref{mis_sipakcell}(c), where most cells are of the "Dyskeratotic" class, and thus an originally "Metaplastic" class image is classified as "Dyskeratotic" class. These wrong predictions might be due to the improper placement of these images in the classes while creating the dataset.

\autoref{mis_sipakcell} shows some misclassified samples from the SIPaKMeD SCI dataset. The possible reasons for the misclassifications are the quality of the image resulting in unclearly visible nuclei like in \autoref{mis_sipakcell}(a) and (b); and the presence of multiple nuclei of cells in the image which is not desired from a single-cell image dataset like in \autoref{mis_sipakcell}(c) and (d).
\section{Statistical Analysis: McNemar's Test}
\begin{table}[]
\caption{Results from McNemar's Test: Null Hypothesis is rejected for every case}
\label{mcnemar}
\resizebox{0.6\columnwidth}{!}{
\begin{tabular}{|c|c|c|c|}
\hline
\textbf{McNemar's Test} & \multicolumn{3}{c|}{\textbf{p-value}} \\ \hline
\textbf{Compared with} &
  \textbf{\begin{tabular}[c]{@{}c@{}}SIPaKMeD \\ SCI\end{tabular}} &
  \textbf{\begin{tabular}[c]{@{}c@{}}Mendeley \\ LBC\end{tabular}} &
  \textbf{\begin{tabular}[c]{@{}c@{}}SIPaKMeD \\ WSI\end{tabular}} \\ \hline
\textbf{Inception v3}     & 0.00012    & 1.88E-08    & 0.0005     \\ \hline
\textbf{DenseNet-161}      & 0          & 0           & 0.0455     \\ \hline
\textbf{ResNet-34}         & 0.0217     & 0.0036      & 0.00433    \\ \hline
\end{tabular}
}
\end{table}

The McNemar's test \cite{dietterich1998approximate} has been performed to justify the viability of the proposed framework with respect to the constituent models in the ensemble. \autoref{mcnemar} shows the results from the test on all the three datasets and clearly, the $p$-value is lower than 5\% in all the cases, and hence the null hypothesis can be rejected, proving that the proposed framework is significantly better than the individual models used to form the ensemble.

\section{Conclusions \& Future Work}\label{conclusions}
In this paper, we propose a fuzzy-fusion based CNN integration method to address the problem of classification of pap-smear based cervical cytology images. The decision scores obtained from CNN classifiers are used as the input of the fuzzy-integral to perform the final classification. With classification accuracies of 99.48\%, 96.33\%, and 98.54\% on Mendeley LBC, SIPaKMeD whole slide images (WSI) and SIPaKMeD single-cell images (SCI) datasets respectively, our proposed method has shown superior performance as compared to other simple fusion methods and has outperformed several existing methods on these datasets. The proposed Sugeno Fuzzy Integral based ensemble is the first such implementation in this domain, and its adaptive weighting system based on the confidence scores of contributing classifiers makes it perform better than the traditional ensemble schemes previously used in the literature as evident from \autoref{comp_ensemble}.

Graph convolution networks (GCN) and attention-gated networks have also shown promising performance in several domains, which engenders our interest to experiment with fuzzy fusion-based methods on these test-beds in the future. The fuzzy measures are selected based on the individual classifier performance on test sets, which is not the optimal solution. Hence, we further plan to implement some evolutionary meta-heuristic optimization algorithm for the selection of the fuzzy measures of the classifiers that might further improve the overall classification performance. We might also incorporate other CNN classifiers to form the ensemble in the future.

\bibliographystyle{unsrtnat}
\bibliography{References.bib}

\end{document}